\pdfoutput=1
\documentclass[sigconf, screen, nonacm]{acmart} 

\usepackage{booktabs}
\usepackage{multirow}
\usepackage{pifont}
\usepackage{algorithm}
\usepackage{algorithmic}
\usepackage{amsmath}
\usepackage{graphicx}
\usepackage{colortbl}
\usepackage{xcolor}
\usepackage{cuted}
\usepackage{caption}
\usepackage{enumitem}

\AtBeginDocument{%
  }

\begin{document}

\title[AIM for VQA Continual Learning]{AIM: Asymmetric Information Masking for \\ Visual Question Answering Continual Learning}



\author{
  \begin{tabular}{@{}c@{}}
    {Peifeng Zhang}$^{1}$, {Zice Qiu}$^{1}$, {Donghua Yu}$^{1}$, {Shilei Cao}$^{1}$, \\
    {Juepeng Zheng}$^{1,2,}$\textsuperscript{\textdagger}, {Yutong Lu}$^{1,2}$, {Haohuan Fu}$^{2,3}$
  \end{tabular}
}
\affiliation{
  \begin{tabular}{@{}c@{}}
    $^{1}$Sun Yat-Sen University \quad $^{2}$National Supercomputing Center in Shenzhen \quad $^{3}$Tsinghua University  \\
    \textsuperscript{\textdagger}Corresponding author
  \end{tabular}
  \country{}
}

\begin{abstract}
In continual visual question answering (VQA), existing Continual Learning (CL) methods are mostly built for symmetric, unimodal architectures. However, modern Vision-Language Models (VLMs) violate this assumption, as their trainable components are inherently asymmetric. This structural mismatch renders VLMs highly prone to catastrophic forgetting when learning from continuous data streams. Specifically, the asymmetry causes standard global regularization to favor the massive language decoder during optimization, leaving the smaller but critical visual projection layers highly vulnerable to interference. Consequently, this localized degradation leads to a severe loss of compositional reasoning capabilities. To address this, we propose Asymmetric Information Masking (AIM), which balances stability and plasticity by applying targeted masks based on modality-specific sensitivity. Experiments on VQA v2 and GQA under continual VQA settings show that AIM achieves state-of-the-art performance in both Average Performance (AP) and Average Forgetting (AF), while better preserving generalization to novel skill-concept compositions.
\end{abstract}

\keywords{Continual Learning, Vision-Language Models,Catastrophic Forgetting, Visual Question Answering}


\maketitle
\pagestyle{plain}

\section{Introduction}

\begin{figure}[t]
  \centering
  \includegraphics[width=\linewidth]{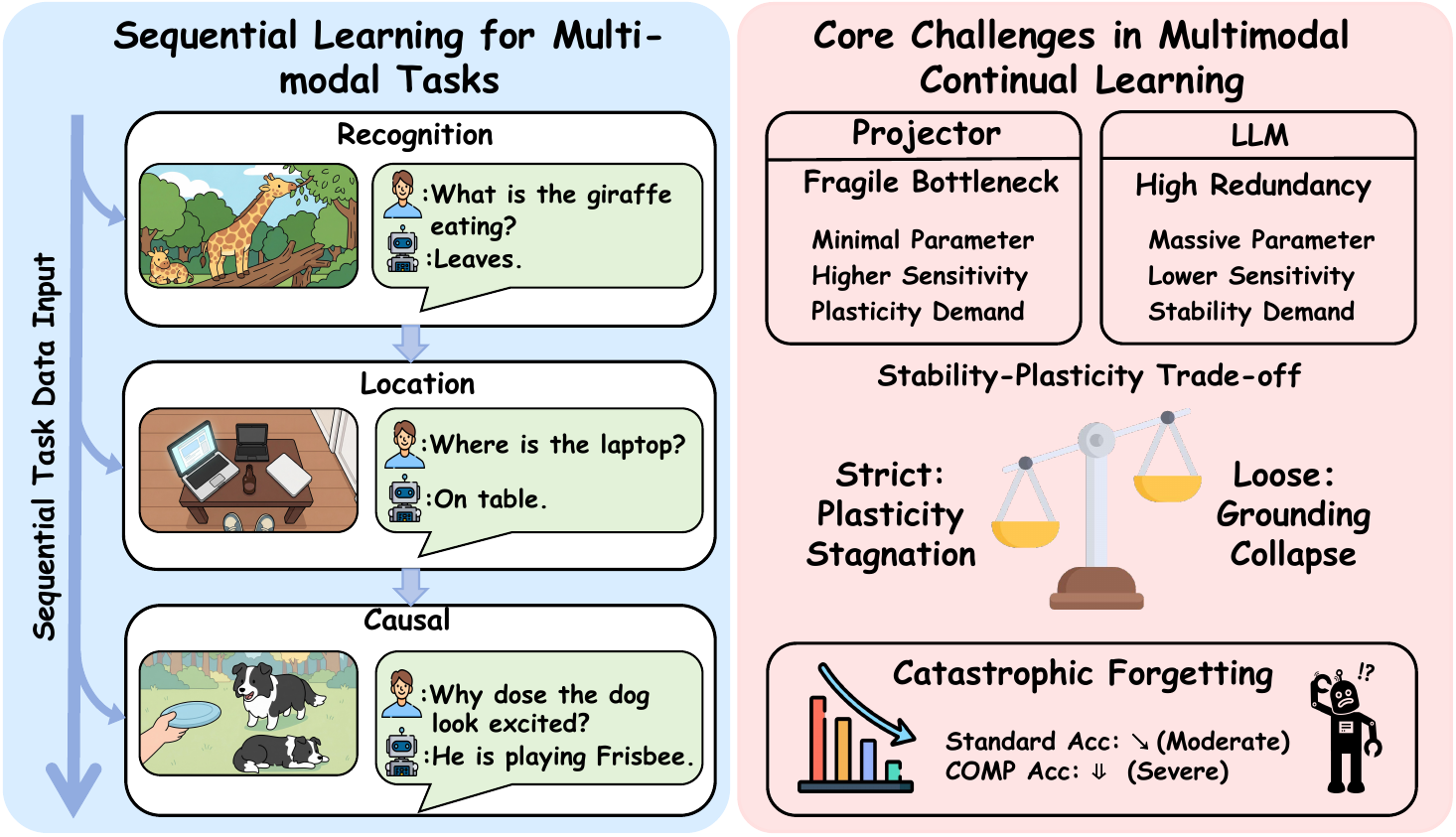} 

  \caption{\textbf{The Asymmetry Dilemma in Multimodal Continual Learning.} 
\textbf{Left:} VLMs in sequential multimodal tasks. 
\textbf{Right:} The structural imbalance between the fragile vision bottleneck and the redundant LLM. Symmetric regularization forces a trade-off between plasticity stagnation and grounding collapse. 
\textbf{Bottom:} The Model faces catastrophic forgetting, where standard accuracy degrades moderately, but compositional generalization suffers a severe collapse.}
\label{fig:dilemma}
\end{figure}

Vision-Language Models (VLMs) \cite{pmlr-v139-radford21a, blip, pmlr-v202-li23q} have established themselves as the foundation of modern multimedia systems by unifying cross-modal perception with linguistic reasoning \cite{alayrac2022flamingovisuallanguagemodel, visual_instruction_tuning}. By leveraging large-scale pre-training, these models achieve state-of-the-art performance in complex tasks such as Visual Question Answering (VQA) \cite{Goyal_2017_CVPR, Yu_2019_CVPR}, cross-modal retrieval \cite{ALBEF}, and video understanding \cite{Yan_Shou_Ge_Wang_Lin_Cai_Tang_2023}. However, deploying VLMs in real-world environments remains challenging, as these models must continuously adapt to non-stationary data streams and learn new visual concepts and reasoning skills over time \cite{zhang2023vqacl}. Directly fine-tuning VLMs on such evolving data inevitably causes \textbf{catastrophic forgetting}, which severely disrupts previously acquired capabilities \cite{french1999catastrophic, mccloskey1989catastrophic}. Continual learning (CL) addresses this issue by enabling models to acquire new knowledge incrementally while preserving previously learned capabilities, thereby avoiding the prohibitive computational cost and privacy concerns associated with retraining from scratch.

To evaluate continual visual question answering more systematically, the {Visual Question Answering Continual Learning} (\textbf{VQACL}) setting has recently been introduced \cite{zhang2023vqacl}.
Unlike standard incremental tasks, VQACL simulates a dual-level data stream in which new visual concepts appear over time and the model must continually acquire new reasoning skills.
As highlighted by recent studies \cite{yang2024quad, zhang2023vqacl,Liu2025ContinualLF}, the primary challenge in such scenarios is balancing \textbf{stability}, which preserves past knowledge, and \textbf{plasticity}, which enables the learning of new information, across both visual and linguistic modalities.
This dual-modality setting requires models not only to retain accuracy on previously learned tasks but also to demonstrate {compositional reasoning} on unseen combinations of reasoning types and visual concepts, such as applying a learned ``counting'' capability to novel object categories.

While various CL strategies have been proposed to mitigate catastrophic forgetting, most of them are developed for symmetric, unimodal architectures and fail to account for the structural asymmetry in VQACL models. Conventional approaches, whether based on regularization \cite{doi:10.1073/pnas.1611835114, 8107520, Aljundi_2018_ECCV} or memory replay \cite{Robins1995CatastrophicFR, 10.5555/3495724.3497059, Wan_2022_CVPR}, implicitly treat all trainable parameters as a {homogeneous set}, assuming that stability and plasticity can be uniformly balanced across the network. Even recent advances tailored for vision-language continual learning \cite{zhang2023vqacl, yang2024quad} inherit this symmetric paradigm. However, we argue that the main limitation of these approaches lies in their failure to account for the {dual asymmetry} between modalities \cite{Wang_2020_CVPR, Peng_2022_CVPR}. In multimodal settings, enforcing a uniform penalty or global gradient constraint often leads to suboptimal adaptation, as different modalities possess distinct optimization dynamics and capacities.

Through detailed analysis (presented in Sec. \ref{sec:observation}), we reveal that this symmetric assumption is explicitly challenged by the structural imbalance inherent in modern VLMs \cite{pmlr-v202-li23q, visual_instruction_tuning} an issue we term the Asymmetry Dilemma (see Fig. \ref{fig:dilemma}). Under prevalent training paradigms, the heavy visual backbone is typically frozen to preserve generic feature extraction. Consequently, incremental updates are largely confined to a lightweight vision-language projector alongside a high-capacity text reasoning module. This disparity in parameter volume induces a systematic \textit{optimization bias}, which often manifests as gradient dominance and disproportionately favors the language decoder \cite{Wang_2020_CVPR, Dai_2024_CVPR}. As a result, the compact visual projector is forced to act as a {fragile semantic bottleneck} \cite{10.5555/3600270.3601550}. As illustrated in Fig. \ref{fig:motivation}, when subjected to uniform CL constraints, even minor parameter shifts within this vulnerable projector can severely disrupt cross-modal grounding. This structural vulnerability not only degrades standard task performance but also largely obscures the catastrophic loss of compositional reasoning capabilities.

Instead of applying a uniform penalty that inherently favors the massive language decoder, we propose Asymmetric Information Masking (AIM) to explicitly decouple the plasticity-stability constraints across the visual, shared, and textual subspaces. By assigning modality-specific masking ratios guided by parameter sensitivity, our framework establishes critical structural anchors within the reasoning core. This strategic isolation forces the remaining free parameters to adapt through compensatory updates, effectively protecting the fragile visual projector from gradient interference. Furthermore, AIM pairs seamlessly with lightweight episodic memory to preserve historical cross-modal alignments.


In summary, our main contributions are as follows:
\begin{itemize}
    \item We analyze the structural asymmetry of trainable components in the VQACL setting and show that symmetric continual learning methods induce a systematic ``gradient dominance'' effect. This bias disproportionately harms the visual projector and obscures the severe degradation of compositional reasoning.
    
    \item We propose Asymmetric Information Masking (AIM), a modality-specific masking method for continual visual question answering. By applying asymmetric parameter constraints across the visual, shared, and textual subspaces, AIM balances stability and plasticity under structurally asymmetric optimization dynamics.
    
    \item Extensive experiments on the VQA v2 \cite{Goyal_2017_CVPR} and GQA \cite{8953451} benchmarks demonstrate that AIM consistently outperforms state-of-the-art methods under identical memory constraints, reducing catastrophic forgetting while also improving compositional generalization in the VQACL setting.
\end{itemize}

\section{Related Work}

\subsection{Continual Learning}

Continual learning studies how models learn from sequential data without forgetting previously acquired knowledge, where catastrophic forgetting remains the central challenge \cite{wang2024survey,mccloskey1989catastrophic}. Existing work mainly reduces forgetting through regularization-based and replay-based strategies \cite{wang2024survey}. Representative regularization methods include EWC \cite{doi:10.1073/pnas.1611835114} and LwF \cite{8107520}, while replay-based methods such as ER \cite{Robins1995CatastrophicFR} and DER \cite{10.5555/3495724.3497059} revisit past samples through memory buffers. Other studies also improve continual learning from the representation side \cite{10.5555/3295222.3295393,rebuffi2017icarl,cha2021co2l,fini2022self}. These methods have shown strong performance in standard unimodal tasks because they either preserve important parameters or reintroduce past data in training.

Yet, most existing studies are developed for settings where the trainable space is more homogeneous, and they do not consider the strong structural imbalance that appears in modern VLMs.

\subsection{Vision-Language Models (VLMs)}

Vision-language models learn aligned visual and textual representations from large-scale paired data and have become a common foundation for multimodal understanding \cite{pmlr-v139-radford21a,ALBEF,blip,pmlr-v202-li23q}. Representative models such as CLIP, ALBEF, BLIP, and BLIP-2 improve cross-modal representation learning and transfer across a wide range of vision-language tasks \cite{pmlr-v139-radford21a,ALBEF,blip,pmlr-v202-li23q}. In many modern VLMs, the visual encoder is kept frozen, while lightweight modules are introduced to connect visual features with language-side reasoning \cite{pmlr-v202-li23q}. This design makes adaptation more efficient and preserves general visual priors learned during pre-training. It also means that most trainable parameters are concentrated in the projector and language-side modules rather than the visual backbone itself.

However, this efficient training paradigm also makes continual adaptation harder, because the visual and language parts no longer update under comparable conditions.

\subsection{Continual Learning in VLMs}

Continual learning in VLMs requires the model to preserve not only previous task performance but also cross-modal alignment and compositional generalization during sequential updates \cite{liu2025vlmsurvey,pmlr-v139-radford21a,gu2021open,du2022learning}. Recent studies have explored continual learning in VQA and related vision-language tasks, showing that this setting is more difficult than standard unimodal continual learning because both visual and linguistic knowledge must be retained across tasks. Existing methods in this line mainly improve continual adaptation through replay, distillation, or other multimodal designs \cite{liu2025vlmsurvey,zhou2022coop,gao2024clipadapter,zhang2022tipadapter,singh2021flava}. For example, VQACL uses prototype-based replay to support continual visual question answering, while QUAD improves memory efficiency through question-only rehearsal \cite{zhang2023vqacl,yang2024quad,liu2025vlmsurvey}. These methods improve knowledge retention in continual VQA and further show that multimodal continual learning often requires more task-specific designs than conventional CL.

Still, these methods focus more on storing, replaying, or constraining historical information than on how different parts of a modern VLM evolve during continual adaptation. As a result, the mismatch between lightweight visual grounding modules and much larger language-side parameters remains insufficiently addressed in existing VQACL methods \cite{liu2025vlmsurvey}. Our method instead focuses on this structural asymmetry in the VQACL setting.

\section{Preliminary}
\label{sec:preliminary}


We begin defining the Visual Question Answering Continual Learning (VQACL) task setting, followed by a characterization of the compositional evaluation protocols throughout this entire research.

\subsection{Problem Formulation}
In standard VQA-CL, a model learns from a non-stationary sequence of tasks $\mathcal{T} = \{T_1, T_2, \dots, T_N\}$. Each task $T_i$ is associated with a training dataset $D_i = \{(v_j, q_j, y_j)\}_{j=1}^{|D_i|}$, where $v_j$ denotes an image, $q_j$ a question, and $y_j$ the corresponding ground-truth answer. The model, parameterized by $\Theta$, is trained to estimate the conditional probability $P_\Theta(y | v, q)$ in an incremental manner. When learning task $T_i$, it has access only to $D_i$ (and optionally a strictly memory-constrained replay buffer $\mathcal{M}$ containing samples from earlier tasks). The main difficulty is to acquire new multimodal knowledge while preserving performance on previously learned tasks $T_{1 \dots i-1}$.

\subsection{Compositional Generalization Setting}

Following the established VQACL protocol \cite{zhang2023vqacl}, we evaluate both long-term knowledge retention and robust compositional generalization. In this specific setting, \textit{compositionality} refers to the model's fundamental ability to accurately interpret novel combinations of previously acquired reasoning skills and diverse visual concepts.

To test this capability, the training sequence is constructed to systematically decouple skills from specific concepts. For instance, the model may acquire the visual concept of a ``truck'' during an earlier task focused on attribute recognition (e.g., answering \textit{``What color is the truck?''}). Subsequently, the model learns the ``counting'' skill using various other objects, while all instances involving counting trucks are strictly excluded from the training data. At evaluation time, the model is assessed under two distinct settings:

\begin{itemize}
    \item \textbf{Standard Testing (Non-COMP):} Evaluates the model on skill-concept combinations following the same data distribution as the training set, measuring the retention of established knowledge.
    \item \textbf{Novel Composition Testing (COMP):} Evaluates the model's zero-shot ability to apply a previously acquired reasoning skill (e.g., \textit{``counting''}) to a visual concept (e.g., \textit{``trucks''}) that was never encountered in that specific combination during training.
\end{itemize}

This distinction is critical because high performance on seen combinations does not guarantee strong compositional generalization. A model might rely on memorizing specific training correlations while remaining incapable of dynamically applying a learned reasoning skill to a separately learned visual concept.

\begin{figure*}[t]
  \centering
  \includegraphics[width=\textwidth]{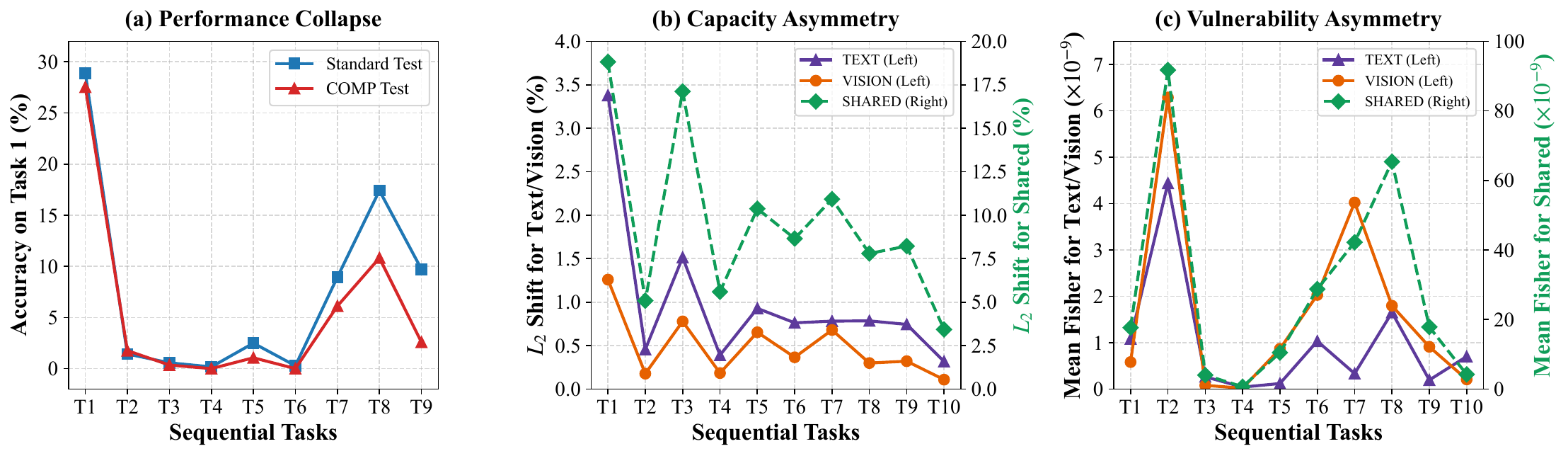}
  \caption{Empirical analysis of multimodal continual learning. \textbf{(a)} Comparison of standard accuracy and compositional reasoning (COMP) performance across sequential tasks. \textbf{(b)} $L_2$ parameter shifts measured across the vision, shared, and text subspaces. \textbf{(c)} Mean Fisher information (parameter sensitivity) evaluated for each corresponding subspace.}
  \label{fig:motivation}
  \Description{A three-panel figure showing standard vs COMP accuracy, L2 parameter shifts, and Fisher sensitivity.}
\end{figure*}

\section{Understanding Dual Asymmetry in Multimodal Networks}
\label{sec:observation}


To understand why multimodal models are prone to catastrophic forgetting~\cite{mccloskey1989catastrophic}, we conducted evaluation experiments using a standard sequential fine-tuning (Vanilla FT) baseline. By training VL-T5~\cite{cho2021unifyingvisionandlanguagetaskstext} (the architecture of which is depicted in Fig.~\ref{fig:figure3}) with a frozen Faster R-CNN~\cite{ren2016fasterrcnnrealtimeobject} visual backbone and monitoring both performance degradation and internal parameter updates, we uncovered the fundamental structural error of applying uniform, symmetric regularization to these multimodal architectures.

\subsection{Architectural Partitioning}
\label{subsec:deconstruction}

Following the common training paradigm for modern Vision Language Models (VLMs)~\cite{li2023blip2bootstrappinglanguageimagepretraining, liu2023visualinstructiontuning}, we kept the heavy visual backbone frozen to preserve the general feature extraction capabilities acquired through pre-training. Although this approach prevents the destruction of established visual representations and reduces computational overhead, it also limits the learnable visual parameters to a small fraction of the entire network. As a result, the continual learning process updates only the lightweight vision-language projector and the large text reasoning modules. Since these components differ significantly in both parameter scale and functional purpose, applying a unified learning strategy inevitably exacerbates the plasticity-stability dilemma~\cite{mccloskey1989catastrophic, doi:10.1073/pnas.1611835114}. To analyze their asymmetric behavior during sequential adaptation in greater detail, we divide the trainable parameter space into three distinct subspaces:

\begin{itemize}
    \item \textbf{Visual Projection Subspace ($\Theta_{vis}$):} Comprising the lightweight region embedding and visual projection layers. With the heavy visual backbone frozen, this remaining subspace acts as a critical ``information bottleneck'' that maps raw visual region features into the shared semantic dimension.
    
    \item \textbf{Cross-modal Alignment Subspace ($\Theta_{shared}$):} Consisting of the shallow-to-intermediate layers of the multimodal encoder (specifically, layers 0--7 in our VL-T5 backbone). Probing studies~\cite{li2020what, Cao2020BehindTS} suggest that these layers primarily handle implicit cross-modal grounding and integrate novel concepts by dynamically adapting visual regions and linguistic tokens.
    
    \item \textbf{Cognitive Reasoning Subspace ($\Theta_{text}$):} Comprising the text embedding layer, the deep encoder layers (layers 8--11), the entire autoregressive decoder, and the language prediction head. As demonstrated by linguistic probing~\cite{48153, Jawahar2019WhatDB}, deep Transformer layers synthesize aligned representations into high-level semantic abstractions. Notably, because the text embeddings and the language head share parameters in the T5 architecture (weight-tying), they must be jointly protected. This subspace functions as the primary reasoning engine and prevents catastrophic concept drift from the input stage by relying on extensive language priors to perform complex compositional reasoning.

\end{itemize}

Mathematically, the trainable parameter space is defined as $\Theta = \Theta_{vis} \cup \Theta_{shared} \cup \Theta_{text}$, providing a structural foundation for our subsequent plasticity analysis.

\subsection{The Performance-Generalization Gap}
\label{subsec:illusion}



Existing CL literature typically relies on standard task performance as the primary metric for evaluating knowledge retention.However, multimodal models often exploit unimodal biases (e.g., language priors) to inflate this metric without genuine cross-modal understanding~\cite{Goyal_2017_CVPR, Agrawal_2018_CVPR}.Relying on standard evaluation is insufficient for continual VQA, as it masks the severity of catastrophic forgetting.



As illustrated in Fig.~\ref{fig:motivation}(a), we track the performance on the initial task ($T_1$) throughout the entire learning sequence. While both metrics undergo significant degradation in the middle stages, their recovery trajectories reveal a striking divergence. At $T_8$, both metrics exhibit a noticeable rebound, with the Standard accuracy reaching 17.41\% and COMP recovering to 10.82\%. However, at the final state ($T_9$), while the Standard accuracy maintains a residual 9.72\%, the COMP accuracy collapses again to a mere 2.59\%. Most importantly, despite temporary fluctuations, the performance gap between the Standard and COMP evaluations consistently widens.

This mismatch reveals a critical illusion of multimodal retention in CL. Since standard test sets follow the same statistical distribution as the training data, the over-parameterized language decoder can maintain or even boost standard performance by leaning on unimodal language priors, for example, by using frequent question-answer shortcuts instead of grounding predictions in visual evidence~\cite{Agrawal_2018_CVPR}. By contrast, the COMP evaluation focuses on unseen combinations, which forces the model to rely on genuine cross-modal understanding. The sharp drop in COMP performance therefore indicates that the underlying visual-linguistic grounding has been substantially damaged. This result suggests that standard metrics alone can hide the real severity of catastrophic forgetting in the visual components of multimodal networks.

\subsection{Analyzing the Dual Asymmetry}
\label{subsec:asymmetry}


Building upon this architectural partition, we diagnose the learning dynamics and forgetting behaviors across the identified subspaces. Specifically, we quantify parameter drift via the $L_2$ norm of updates ($\Delta \Theta$) and measure parameter importance using the Mean Fisher Information Matrix~\cite{doi:10.1073/pnas.1611835114, JMLR:v21:17-678} to characterize the stability-plasticity trade-off of each specific modality.


\paragraph{Capacity Asymmetry (Parameter Volume and Shift)} 
As illustrated in Fig.~\ref{fig:motivation}(b), we observe a strict hierarchy in update magnitudes: $\Delta\Theta_{shared} \gg \Delta\Theta_{text} > \Delta\Theta_{vis}$. Over 10 tasks, $\Theta_{shared}$ exhibits an average shift of 9.60\% to actively establish novel alignments. However, the parameter count remains extremely imbalanced; text modules contain 100$\times$ more parameters than the vision projector. Consequently, the text module's sheer volume dominates the global optimization objective, leaving the sensitive vision bottleneck starved of the gradient updates essential for maintaining robust alignment.

\paragraph{Vulnerability Asymmetry (Parameter Sensitivity)}
Fig.~\ref{fig:motivation}(c) evaluates parameter sensitivity. As expected, $\Theta_{shared}$ shows the highest mean sensitivity ($28.3 \times 10^{-9}$) due to its direct role in bridging multimodal inputs. However, a critical anomaly exists between the remaining two modalities: despite undergoing the smallest parameter shift, $\Theta_{vis}$ exhibits a mean sensitivity ($1.68 \times 10^{-9}$) that significantly exceeds the large-scale $\Theta_{text}$ module ($0.98 \times 10^{-9}$). This establishes $\Theta_{vis}$ as a highly sensitive semantic anchor. Even minor deviations in this low-capacity bottleneck can fundamentally disrupt the visual-language grounding established during pre-training.

\paragraph{The Structural Flaw of Symmetric Regularization}
Traditional regularization methods (e.g., EWC~\cite{doi:10.1073/pnas.1611835114}, MAS~\cite{Aljundi_2018_ECCV}) compute a global penalty by summing the weighted parameter shifts:


\begin{equation}
\mathcal{L}_{reg}(\Theta)=\lambda \sum_{k \in \{vis,shared,text\}}
\left[
\sum_{\theta_i \in \Theta_k} I_i(\theta_i-\theta_{i,\mathrm{old}})^2
\right],
\end{equation}

where $I_i$ denotes the estimated parameter importance (e.g., Fisher Information). When applied to MLLMs, this symmetric paradigm leads to the gradient dominance of the language decoder. Because $\Theta_{text}$ contains exponentially more parameters than $\Theta_{vis}$ ($|\Theta_{text}| \gg |\Theta_{vis}|$), the summation over the text module dominates the global loss formulation. The aggregate penalty for the text layers vastly outweighs that of the vision module, nullifying the fact that individual visual parameters possess higher average importance.

This structural bias forces gradient-based optimizers to disproportionately allocate the regularization budget to the text modality, leaving the sensitive visual bottleneck under-regularized. This observation confirms the inadequacy of global, symmetric regularization for asymmetric architectures and provides the explicit motivation for our proposed AIM framework, which counteracts this budget misallocation via targeted architectural masking.

\begin{figure*}[t]
  \centering
  \includegraphics[width=\textwidth]{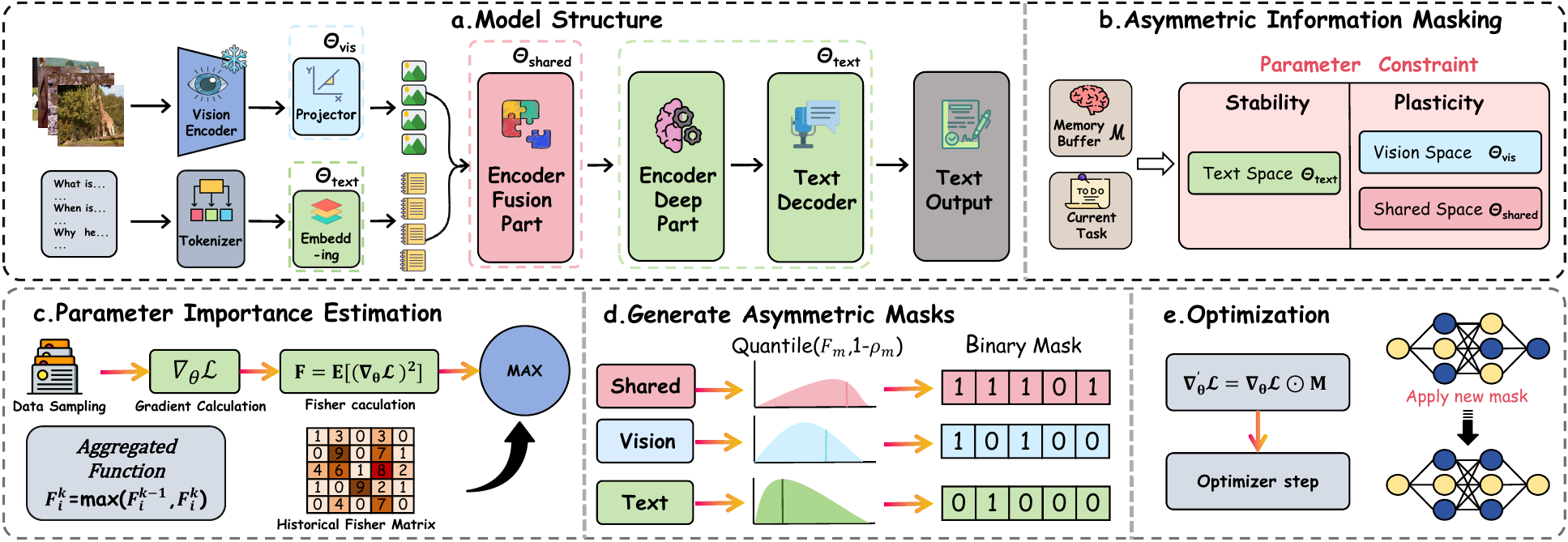}
  \caption{\textbf{Overview of the Asymmetric Information Masking (AIM) framework.} \textbf{(a) Model Structure:} The VLM architecture is explicitly partitioned into three modality-specific subspaces based on their functional roles: Vision (cyan blocks), Shared (pink blocks), and Text (green blocks). \textbf{(b) AIM:} AIM integrates asymmetric modality constraints with a memory buffer. \textbf{(c) Importance Estimation:} Diagonal Fisher information aggregates parameter sensitivity. \textbf{(d) Generate Masks:} Modality-specific thresholds generate binary masks to decouple optimization dynamics. \textbf{(e) Optimization:} Masked gradients to update the model.}
  \label{fig:figure3}
  \Description{A three-part diagram showing the AIM framework: Part A shows the calculation of Fisher information and its aggregation via a Max function. Part B illustrates the partitioning of parameters into Shared, Vision, and Text subspaces, each with a different quantile-based threshold and binary mask. Text has the most restrictive mask for stability, while Vision and Shared are more plastic. Part C shows the application of these binary masks to gradients during the optimizer step to update the neural network.}
\end{figure*}

\section{Methodology}
\label{sec:method}

Motivated by the inherent structural asymmetry and the gradient dominance of the language decoder identified in Sec.~\ref{sec:observation}, we propose  \textbf{A}symmetric \textbf{I}nformation \textbf{M}asking (AIM) for continual visual question answering. Rather than applying a uniform global penalty that biases optimization, our framework explicitly decouples the plasticity constraints across the visual, shared, and textual subspaces. By assigning tailored masking ratios guided by empirical sensitivity profiles, AIM effectively isolates the fragile visual projection bottleneck from catastrophic misalignment while preserving the reasoning capacity for zero-shot compositional generalization.  A comprehensive visual overview of the AIM framework is provided in Fig. \ref{fig:figure3}, which illustrates its three main phases: Parameter Importance Estimation, Generate Asymmetric Masks, and Optimization.

\subsection{Parameter Importance Estimation}
\label{subsec:fisher}
To determine which parameters are crucial for previously learned tasks, we estimate their importance using the empirical Fisher Information Matrix (FIM) \cite{JMLR:v21:17-678}, a foundational metric in weight consolidation literature \cite{doi:10.1073/pnas.1611835114}. For the network parameters $\theta$, we compute the diagonal elements of the FIM to evaluate parameter sensitivity at a fine-grained level. Specifically, after learning a task $T_k$, we sample a small subset of the training data $\mathcal{D}_k$. Let $\mathcal{L}(\theta)$ be the standard cross-entropy loss. The Fisher information $F_{k,i}$ for the $i$-th parameter $\theta_i$ is approximated by the squared gradients:
\begin{equation}
    F_{k,i} = \mathbb{E}_{(v, q, a) \sim \mathcal{D}_k} \left[ \left( \frac{\partial \mathcal{L}(v, q, a; \theta)}{\partial \theta_i} \right)^2 \right]
\end{equation}

In continual learning, standard additive aggregation of Fisher information \cite{doi:10.1073/pnas.1611835114} monotonically accumulates penalty weights, which over-constrains the parameter space and limits plasticity \cite{pmlr-v80-schwarz18a}. To mitigate this, we propose a recursive maximum-based aggregation function. Let $F_{i}^{(k)}$ denote the aggregated importance after task $k$:
\begin{equation}
    F_{i}^{(k)} = \max\left(F_{i}^{(k-1)}, F_{k, i}\right)
\end{equation}
This non-additive operation bounds the accumulated sensitivities. It ensures that parameters crucial to past tasks remain protected without sacrificing the network capacity required for subsequent visual question answering.



\begin{table*}[t]
  \centering
  
  \caption{Continual learning performance on VQA v2 and GQA. `Standard' measures the retention of statistical mappings, while `COMP' evaluates zero-shot compositional generalization. Results are reported in percentage (\%). Bold and underlined values indicate the best and the second-best results, respectively. `\#Mem.' denotes the episodic memory size.}
  \label{tab:main_results}
  \setlength{\tabcolsep}{5pt}
  \begin{tabular*}{\textwidth}{@{\extracolsep{\fill}}l ccccc ccc}
    \toprule
    \multirow{3}{*}{\textbf{Method}} & \multicolumn{5}{c}{\textbf{VQA v2 (Linguistic Incremental)}} & \multicolumn{3}{c}{\textbf{GQA (Scene Incremental)}} \\
    \cmidrule(lr){2-6} \cmidrule(lr){7-9}
    & \multirow{2}{*}{\textbf{\#Mem.}} & \multicolumn{2}{c}{\textbf{Standard Test}} & \multicolumn{2}{c}{\textbf{Novel Composition}} & \multirow{2}{*}{\textbf{\#Mem.}} & \multicolumn{2}{c}{\textbf{Standard Test}} \\
    \cmidrule(lr){3-4} \cmidrule(lr){5-6} \cmidrule(lr){8-9}
    & & AP ($\uparrow$) & AF ($\downarrow$) & AP ($\uparrow$) & AF ($\downarrow$) & & AP ($\uparrow$) & AF ($\downarrow$) \\
    \midrule
    Joint & - & 51.64 & - & 51.10 & - & - & 40.65 & - \\
    Vanilla & None & 14.49 & 30.80 & 11.79 & 27.16 & None & 23.83 & 20.49 \\
    \midrule
    EWC \cite{doi:10.1073/pnas.1611835114} & None & 15.77 & 30.62 & 12.83 & 28.16 & None & 23.28 & 20.31 \\
    LwF \cite{8107520} & None & 22.34 & 23.00 & 16.65 & 28.25 & None & 22.07 &  20.17 \\
    MAS \cite{Aljundi_2018_ECCV} & None & 20.56 & 11.16 & 23.90 & 6.24 & None & 19.46 & 14.93 \\
    ER \cite{Robins1995CatastrophicFR} & 5000 & 36.99 & 5.99 & 33.78 & 5.76 & 500 & 36.43 & \underline{6.18}   \\
    DER \cite{buzzega2020darkexperiencegeneralcontinual} & 5000 & 35.35 & 8.62 & 31.52 & 8.59 & 500 & \underline{36.46} & 6.96 \\
    \midrule
    VQACL \cite{zhang2023vqacl} & 5000 & 38.77 & \underline{3.96} & 35.40 & 4.90 & 500 & 35.26 & 7.37 \\
    QUAD \cite{marouf2025askrememberquestionsonlyreplay} & 5000 & \underline{39.25} & 4.91 & \underline{40.00} & \underline{3.81} & 500 & 30.14 & 6.94 \\
    \midrule
    \textbf{AIM (Ours)} & 5000 & \textbf{43.35} & \textbf{1.56} & \textbf{42.03} & \textbf{3.35} & 500 & \textbf{37.51} & \textbf{5.53} \\
    \bottomrule
  \end{tabular*}
\end{table*}

\subsection{Generate Asymmetric Masks}
\label{subsec:aims}
Traditional parameter-isolation methods typically apply a globally uniform masking threshold across the entire architecture. However, our architectural analysis (Sec.~\ref{subsec:deconstruction}) firmly establishes that the parameter spaces $\Theta_{vis}$, $\Theta_{shared}$, and $\Theta_{text}$ possess fundamentally distinct capacities and optimization dynamics.

To address this disparity, AIM introduces modality-specific protection ratios. Let $\rho_m \in [0, 1]$ denote the masking ratio for a specific subspace $m \in \{vis, shared, text\}$. This ratio dictates the proportion of the most sensitive parameters within that subspace that must be strictly frozen. For each subspace $m$, we collect its corresponding Fisher values $\mathbf{F}_m = \{F_{i}^{(k)} \mid \theta_i \in \Theta_m\}$. We then compute a dynamic, subspace-specific threshold $\tau_m$ by finding the $(1 - \rho_m)$-th quantile of the Fisher distribution within that specific subspace:
\begin{equation}
    \tau_m = \text{Quantile}(\mathbf{F}_m, 1 - \rho_m)
\end{equation}
This formulation ensures that exactly $\rho_m \times 100\%$ of the parameters in $\Theta_m$ have a Fisher sensitivity greater than or equal to $\tau_m$. Based on these asymmetric thresholds, we generate a binary gradient mask $M_i$ for each parameter $\theta_i \in \Theta_m$:
\begin{equation}
    M_i = 
    \begin{cases} 
      0 & \text{if } F_{i}^{(k)} \geq \tau_m \text{ (Crucial, Frozen)} \\
      1 & \text{if } F_{i}^{(k)} < \tau_m \text{ (Plastic, Trainable)}
    \end{cases}
\end{equation}

This asymmetric configuration mitigates the structural optimization bias at the architectural level. Specifically, we enforce a strict $\rho_{\text{text}}$ to constrain the parameter updates within the text decoder. Concurrently, we maintain a relaxed $\rho_{\text{vis}}$ to preserve the plasticity of the vision module, enabling it to encode novel visual distributions.

\subsection{Optimization}
\label{subsec:optimization}
During the backward pass for a new task $T_{k+1}$, the generated masks are applied directly to the network gradients. Let $g_i = \frac{\partial \mathcal{L}}{\partial \theta_i}$ denote the raw gradient. The effective gradient $g'_i$ for the optimization step is computed via the Hadamard product:
\begin{equation}
    g'_i = g_i \odot M_i
\end{equation}
While this structural masking effectively mitigates inter-task interference, it remains orthogonal to standard rehearsal techniques. By coupling this gradient modulation with a lightweight episodic memory buffer, our framework preserves established cross-modal alignments while accommodating novel concepts throughout the continual learning process.

\section{Experiments}

\subsection{Experimental Setup}
\label{subsec:exp_setup}

\paragraph{Datasets and Protocols}
We evaluate our method on two visual question answering benchmarks under continual learning protocols. (1) VQA v2 (Linguistic-Incremental): Following the VQACL protocol~\cite{zhang2023vqacl}, the dataset is partitioned into 10 tasks driven by different question types. The COMP split employs a 5-fold object-independent cross-validation protocol (G1-G5), ensuring that all testing instances involve unseen combinations of learned reasoning skills and visual concepts. (2) GQA (Scene-Incremental): Following the CLOVE benchmark protocol~\cite{nguyen2023clove}, this dataset is sequentially partitioned into 6 tasks shifting across distinct environmental scenes (e.g., ShopAndDining, Outdoors). This setup simulates drastic visual domain shifts to stress-test the stability of the visual representation bottleneck. Detailed descriptions of each sub-task and their dataset statistics are provided in Appendix~\ref{supp:datasets}.


\paragraph{Metrics} 
We evaluate the performance using two standard continual learning metrics~\cite{10.5555/3295222.3295393,Chaudhry_2018}: \textbf{Average Performance (AP)} and \textbf{Average Forgetting (AF)}. Let $a_{i,j}$ denote the performance on task $T_i$ after the model has finished training on task $T_j$. 
Upon completing all $N$ tasks, \textbf{AP} is defined as $AP = \frac{1}{N} \sum_{i=1}^{N} a_{i,N}$, which reflects the model's overall proficiency across the entire sequence. 
To quantify catastrophic interference, \textbf{AF} is computed as $AF = \frac{1}{N-1} \sum_{i=1}^{N-1} \max_{j \in \{i, \dots, N-1\}} (a_{i,j} - a_{i,N})$. 
A lower AF value indicates superior knowledge retention and stronger resistance to forgetting.

\paragraph{Baselines}
We compare our proposed AIM against several representative methods: (1) \textit{Regularization-based:} EWC \cite{doi:10.1073/pnas.1611835114}, LwF \cite{8107520}, and MAS \cite{Aljundi_2018_ECCV}; (2) \textit{Rehearsal-based:} Experience Replay (ER) \cite{Robins1995CatastrophicFR} and Dark Experience Replay (DER) \cite{buzzega2020darkexperiencegeneralcontinual}; (3) \textit{Multimodal-specific:} the prototype-based me
method VQACL \cite{zhang2023vqacl} and the question-only replay strategy QUAD \cite{marouf2025askrememberquestionsonlyreplay}. Detailed descriptions for each baseline method are provided in Appendix~\ref{supp:baselines}.

\paragraph{Implementation Details}
We build our framework upon the pre-trained VL-T5 backbone. Following the standard configuration~\cite{zhang2023vqacl}, we use a Faster R-CNN \cite{ren2016fasterrcnnrealtimeobject} trained on Visual Genome \cite{krishna2016visualgenomeconnectinglanguage} to extract 36 region features per image. The transformer backbone stacks 12 blocks for both the encoder and decoder, each with 12 attention heads and an embedding dimension of $d=768$.To demonstrate the generalizability of AIM, we also apply it to modern large vision-language models (LLaVA \cite{liu2023visualinstructiontuning}) under the standard settings detailed in Appendix~\ref{supp:llava}. For fair comparison, all rehearsal methods (including ours) are strictly constrained to a unified episodic memory buffer size: $M=5,000$ samples for VQA v2 and $M=500$ for GQA. Models are trained for 3 epochs per task with a batch size of 80, using the AdamW optimizer 
\cite{loshchilov2019decoupledweightdecayregularization} with an initial learning rate of $10^{-4}$. Our AIM framework applies asymmetric protection ratios ($\rho_{vis}=0.3, \rho_{shared}=0.1, \rho_{text}=0.5$) to balance the distinct stability and plasticity demands. Detailed parameter analysis are provided in Appendix~\ref{supp:parameter_sensitivity}.  Specifically, to maintain computational efficiency during the Fisher sensitivity calculation at the end of each task, we randomly sample $N=500$ instances from the training data. Full hyperparameter configurations and hardware specifications are provided in Appendix~\ref{supp:implementation}.

\begin{table*}[t]
  \centering
  \caption{Fine-grained VQA performance AP (\%) on the Novel and Seen skill-concept compositions of VQA v2. Best results are in \textbf{bold}, and second-best are \underline{underlined}.}
  \label{tab:comp_breakdown}
  \Description{A table showing fine-grained VQA performance with columns for Method, and interleaved Novel and Seen sub-columns for Groups 1 through 5, along with the overall Average.}
  \begin{tabular}{l cccccccccccc}
    \toprule
    \multirow{2}{*}{\textbf{Method}} & \multicolumn{2}{c}{\textbf{Group-1}} & \multicolumn{2}{c}{\textbf{Group-2}} & \multicolumn{2}{c}{\textbf{Group-3}} & \multicolumn{2}{c}{\textbf{Group-4}} & \multicolumn{2}{c}{\textbf{Group-5}} & \multicolumn{2}{c}{\textbf{Avg}} \\
    \cmidrule(lr){2-3} \cmidrule(lr){4-5} \cmidrule(lr){6-7} \cmidrule(lr){8-9} \cmidrule(lr){10-11} \cmidrule(lr){12-13}
    & Novel & Seen & Novel & Seen & Novel & Seen & Novel & Seen & Novel & Seen & Novel & Seen \\
    \midrule
    DER \cite{buzzega2020darkexperiencegeneralcontinual} & 30.80 & 29.89 & 32.19 & 33.24 & 34.88 & 34.08 & 29.60 & 30.90 & 30.14 & 32.56 & 31.52 & 32.13 \\
    ER \cite{Robins1995CatastrophicFR} & 34.52 & 37.03 & 33.40 & 35.55 & 34.79 & 34.20 & 33.86 & 35.02 & 32.34 & 35.91 & 33.78 & 35.54 \\
    VQACL \cite{zhang2023vqacl} & 36.12 & 37.99 & 35.39 & 36.92 & 36.26 & 35.16 & 34.85 & 35.64 & 34.36 & 36.28 & 35.40 & 36.40 \\
    QUAD \cite{marouf2025askrememberquestionsonlyreplay} & \underline{39.19} & \underline{41.06} & \underline{38.40} & \underline{39.50} & \textbf{43.15} & \underline{39.19} & \underline{40.01} & \underline{40.72} & \underline{39.20} & \underline{40.62} & \underline{40.00} & \underline{40.21} \\
    \midrule
    \textbf{AIM (Ours)} & \textbf{41.92} & \textbf{43.66} & \textbf{42.69} & \textbf{44.20} & \underline{42.22} & \textbf{44.13} & \textbf{41.61} & \textbf{44.60} & \textbf{41.70} & \textbf{44.06} & \textbf{42.03} & \textbf{44.13} \\
    \bottomrule
  \end{tabular}
\end{table*}

\subsection{Main Results}
\label{subsec:main_results}
We report the overall continual learning results of AIM against the baselines in Table~\ref{tab:main_results}, and provide a fine-grained analysis of compositional generalization in Table~\ref{tab:comp_breakdown}.



\paragraph{Performance Analysis on Standard Settings.}
As illustrated in the ``Standard Test'' columns of Table~\ref{tab:main_results}, while traditional rehearsal methods like ER~\cite{Robins1995CatastrophicFR} and DER~\cite{10.5555/3495724.3497059} retain previous knowledge better than regularization techniques, they implicitly treat the multimodal network symmetrically. Consequently, they fail to balance the distinct stability and plasticity demands of different modalities, leading to suboptimal adaptation and knowledge degradation (e.g., DER records an 8.62\% AF on VQA v2). Furthermore, although the question-only replay strategy in QUAD~\cite{yang2024quad} reduces memory overhead, discarding historical visual exemplars impairs the model's ability to maintain robust vision-language alignments over time. This limitation becomes especially apparent in the scene-incremental GQA benchmark, where domain shifts heavily rely on visual context, limiting QUAD's performance to 30.14\% AP and resulting in an elevated AF of 6.94\%.

In contrast, AIM resolves these optimization conflicts through an asymmetric subspace isolation strategy. By explicitly applying modality-specific protection masks, our framework prevents the high-capacity language decoder from overwriting the sensitive visual projector. With this targeted protection, AIM achieves the highest AP and the lowest AF across both benchmarks. On the VQA v2 standard test, it outperforms the best-performing baseline, QUAD, by \textbf{+4.10\%} AP (43.35\% vs. 39.25\%) while reducing the AF to 1.56\%. Simultaneously, it achieves state-of-the-art results on GQA, yielding 37.51\% AP and constraining the AF to 5.53\%.




\paragraph{Performance Analysis on Novel Composition Testing.}
Beyond retaining learned distributions, continual VQA models must also generalize to unseen skill-concept combinations. As shown in the ``Novel Comp'' columns of Table~\ref{tab:main_results}, AIM outperforms all continual learning baselines, reaching 42.03\% AP and 3.35\% AF. Furthermore, Table~\ref{tab:comp_breakdown} provides a fine-grained breakdown across five compositional groups on VQA v2, where AIM achieves the highest average performance on both ``Novel'' and ``Seen'' compositions. It ranks first in all five ``Seen'' groups and four ``Novel'' groups. The exception is in Group-3 Novel, where QUAD~\cite{yang2024quad} surpasses AIM (43.15\% vs. 42.22\%). However, when averaged across all groups, AIM exceeds QUAD by \textbf{+2.03\%} AP on Novel compositions (42.03\% vs. 40.00\%) and by \textbf{+3.92\%} AP on Seen configurations (44.13\% vs. 40.21\%).

The compositional evaluation underscores the efficacy of our asymmetric design. Regularization methods exhibit significant degradation in zero-shot settings (e.g., MAS yields 23.90\% AF), while traditional rehearsal approaches (e.g., ER~\cite{Robins1995CatastrophicFR}, DER~\cite{10.5555/3495724.3497059}, and VQACL~\cite{zhang2023vqacl}) plateau in the low-to-mid 30\% AP range with elevated forgetting rates. Conversely, AIM approaches the offline Joint training upper bound (51.10\% AP) and constrains the compositional AF to 3.35\%. This performance validates our core hypothesis: isolating the visual projector from cross-modal gradient interference mitigates the entanglement of visual concepts and reasoning skills. By maintaining disentangled representations, the model can dynamically recombine these elements, facilitating zero-shot compositional generalization.

\subsection{Ablation Study}
\label{subsec:ablation}

\paragraph{Asymmetric Masking.} 
We compare our modality-specific masking ($\rho_{text}=0.5, \rho_{vis}=0.3$) against \textit{Uniform} ($\rho=0.3$ globally) and \textit{Swapped} ($\rho_{text}=0.3, \rho_{vis}=0.5$) variants. Both alternatives lead to performance degradation. Notably, \textit{Swapped Masking} increases Average Forgetting (AF) to 2.59\% on VQA v2. This confirms that over-constraining the visual bottleneck hinders novel concept adaptation, while under-protecting the language decoder exacerbates the forgetting of previously learned reasoning skills.


\paragraph{Fisher Aggregation.}
Replacing our \textit{Max} strategy with summation (\textit{Sum}) alters the stability-plasticity trade-off. On VQA v2, \textit{Sum} yields a lower AF (1.33\% vs. 1.56\%) but restricts AP to 42.96\%. This gap stems from the monotonic accumulation of Fisher penalties under the \textit{Sum} operator, which constrains the parameter space and reduces plasticity. This effect is pronounced in the scene-incremental GQA benchmark, where \textit{Sum} results in a 37.39\% AP and a 5.99\% AF. Conversely, our \textit{Max} operator bounds these penalties, preserving sufficient flexibility for superior overall performance.


\paragraph{Exemplar-Free Setting ($M=0$).} 
To isolate the contribution of asymmetric masking, we evaluate our framework without episodic memory (``AIM w/o Memory'' in Table~\ref{tab:ablation}). In this setting, AIM achieves 26.04\% AP on VQA v2, outperforming traditional regularization baselines (e.g., MAS \cite{Aljundi_2018_ECCV} and LwF \cite{8107520}), which reach a maximum of 22.34\% AP under identical constraints. This demonstrates that structural parameter isolation mitigates catastrophic forgetting independently of data rehearsal.

\begin{figure}[t]
  \centering
  \includegraphics[width=1.0\linewidth]{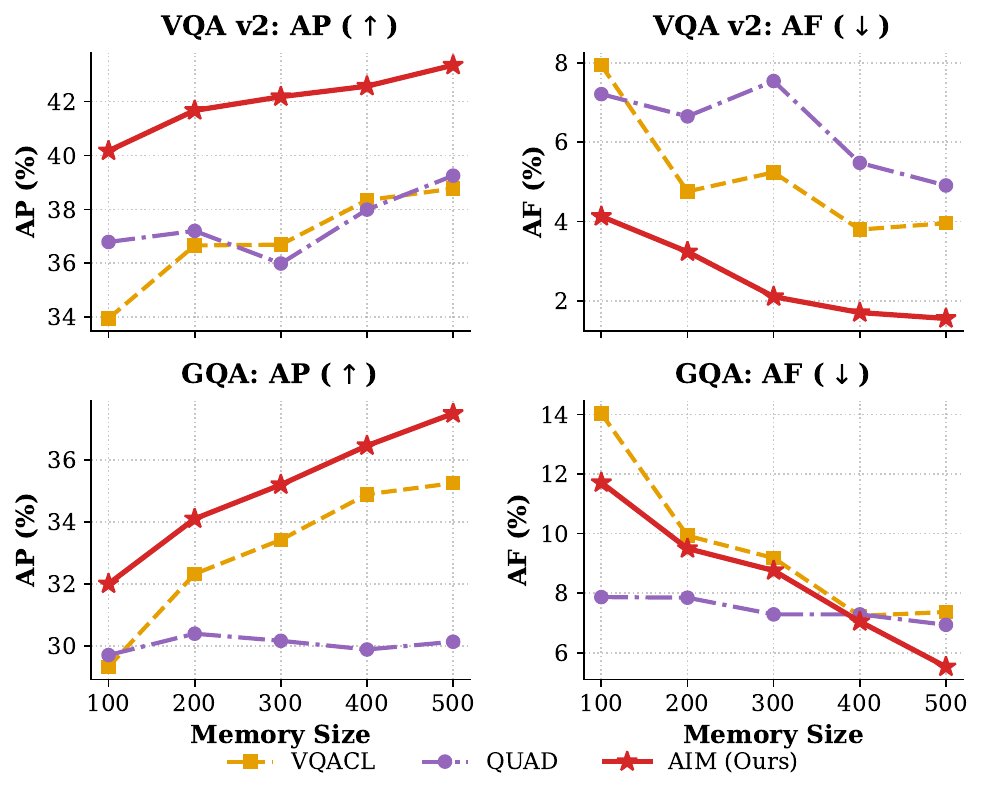}
  \caption{Sensitivity analysis on the memory size ($M$) in terms of Average Performance (AP) and Average Forgetting (AF) ON VQA v2 and GQA.}
  \Description{A 2 by 2 grid of line charts showing the effect of memory size on Average Performance and Average Forgetting for VQA v2 and GQA datasets. The x-axis represents memory size, and the y-axis represents the metric percentage. Across all four charts, the red line representing our AIM method consistently shows higher performance and lower forgetting compared to the baselines, VQACL and QUAD, especially at lower memory capacities.}
  \label{fig:memory_ablation}
\end{figure}

\paragraph{Sensitivity to Memory Size.} 
Fig.~\ref{fig:memory_ablation} illustrates performance across varying memory capacities. While QUAD's AP plateaus on GQA suggesting that text-only buffers struggle to compensate for the missing visual context required for scene adaptation, AIM consistently yields superior AP and lower AF across all memory constraints. This highlights its efficiency under strict memory budgets.

\begin{table}[t]
  \centering
  \caption{Ablation study of AIM components on VQA v2 and GQA. AP ($\uparrow$) and Average Forgetting (AF, $\downarrow$) are reported in percentage (\%).}
  \label{tab:ablation}
  \resizebox{\columnwidth}{!}{
  \begin{tabular}{l | c | cc | cc}
    \toprule
    \multirow{2}{*}{\textbf{Method}} & \multirow{2}{*}{\textbf{Agg.}} & \multicolumn{2}{c|}{\textbf{VQA v2 ($M=5000$)}} & \multicolumn{2}{c}{\textbf{GQA ($M=500$)}} \\
    \cmidrule{3-4} \cmidrule{5-6}
    & & \textbf{AP($\uparrow$)} & \textbf{AF($\downarrow$)} & \textbf{AP($\uparrow$)} & \textbf{AF($\downarrow$)} \\
    \midrule
    Uniform Masking & Max & 40.46 & 1.76 & 36.77 & 5.99 \\
    Swapped Masking & Max & 42.61 & 2.59 & 36.96 & 6.27 \\
    \midrule
    AIM w/o Memory & Max & 26.04 & 19.99 & 24.07 & 21.15 \\
    AIM (Sum Agg.) & Sum & 42.96 & \textbf{1.33} & 37.39 & 5.99 \\
    AIM (Ours) & Max & \textbf{43.35} & \underline{1.56} & \textbf{37.51} & \textbf{5.53} \\
    \bottomrule
  \end{tabular}
  }
\end{table}

\subsection{Further Analysis}

\begin{figure}[t]
  \centering
  \includegraphics[width=\linewidth]{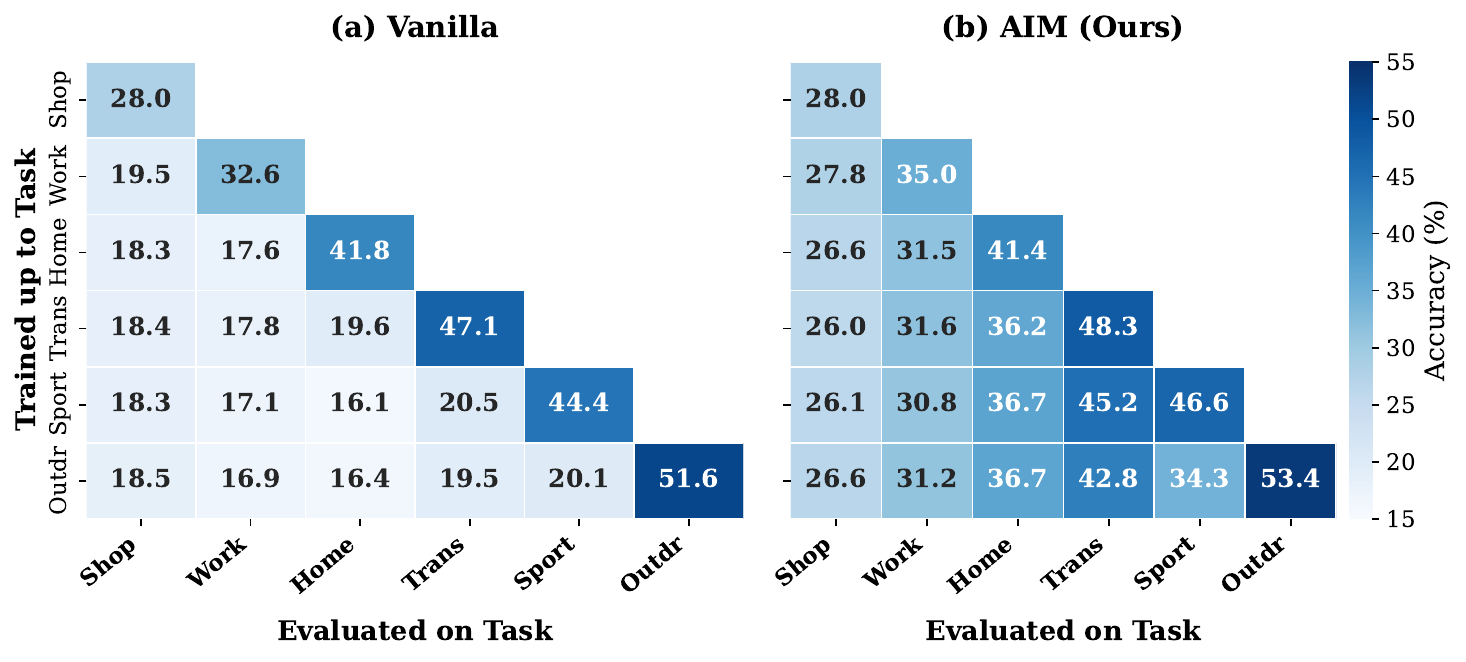}
  \caption{Task-by-task evaluation matrices on the GQA benchmark. \textbf{(a)} \textit{Vanilla} baseline, illustrating performance degradation on previously learned scenes. \textbf{(b)} \textit{AIM (Ours)}, demonstrating performance retention across sequential tasks.}
  \Description{Two side-by-side heatmaps comparing the task-by-task evaluation matrices of the Vanilla model and the AIM model on the GQA benchmark. The left heatmap for the Vanilla model shows bright colors on the diagonal but fading colors in the vertical columns, indicating severe forgetting of earlier tasks. The right heatmap for AIM shows consistent, darker colors across the columns, demonstrating strong retention of prior knowledge as new tasks are learned.}
  \label{fig:heatmap}
\end{figure}

\paragraph{Analysis Stability and Plasticity.}
To examine the underlying learning dynamics, we visualize the task-by-task evaluation matrices on the GQA benchmark (Fig.~\ref{fig:heatmap}). Each cell $(i, j)$ denotes the accuracy on $T_j$ after training up to $T_i$. Without structural protection, the \textit{Vanilla} baseline (Fig.~\ref{fig:heatmap}a) exhibits catastrophic forgetting, characterized by column-wise performance decay as the domain shifts. For instance, accuracy on $T_1$ (\textit{Shop}) falls from 27.97\% to 18.53\%, and $T_4$ (\textit{Trans}) decreases from 47.10\% to 19.47\%. Conversely, AIM (Fig.~\ref{fig:heatmap}b) demonstrates substantial stability; $T_1$ accuracy is maintained at 26.60\% even after five subsequent tasks. Furthermore, AIM consistently achieves higher diagonal values (e.g., 53.4\% vs. 51.6\% on $T_6$), suggesting that isolating modality-specific gradients secures prior knowledge without compromising the plasticity required for novel distributions. This comparison validates that our asymmetric design mitigates inter-task interference, facilitating both knowledge retention and rapid adaptation.

\begin{figure}[t]
  \centering
  \includegraphics[width=1.0\columnwidth]{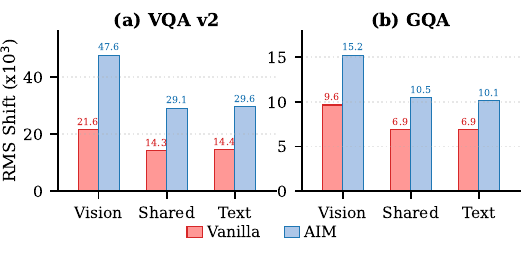}
  \caption{Comparison of RMS parameter shifts across the Vision, Shared, and Text subspaces for the \textit{Vanilla} baseline and AIM on the VQA v2 and GQA benchmarks.}
  \Description{A bar chart comparing the RMS parameter shift across Vision, Shared, and Text subspaces on VQA v2 and GQA. The Vision subspace shows the highest shift, which increases under AIM, while the Text subspace remains the most stable.}
  \label{fig:param_shift}
\end{figure}

\paragraph{Analysis of Parameter Dynamics.}
We employ Root Mean Square (RMS) shift ($\times 10^3$) to quantify weight trajectories, preventing the language decoder's large parameter volume from masking the vision module's optimization dynamics. As shown in Fig.~\ref{fig:param_shift}, the vision projector exhibits the largest relative shift, identifying it as the primary locus for adapting to novel visual distributions. While the language core accounts for most total updates, the projector sustains higher per-parameter variance. Standard fine-tuning yields a constrained shift (21.57 on VQA v2), where unprotected perturbations can disrupt multimodal alignments. By freezing sensitive parameters, AIM necessitates a larger gradient flux through the remaining weights. Consequently, the vision shift increases to 47.56, indicating that the model concentrates its plasticity on parameters that do not compromise prior knowledge.

\section{Conclusion}

In this paper, we study continual visual question answering and show that the key difficulty lies not only in catastrophic forgetting, but also in the asymmetric update dynamics of multimodal models. Standard evaluation can overestimate retention by hiding a larger drop in compositional reasoning, while symmetric constraints often fail to protect the fragile visual bottleneck. To address this issue, we propose AIM, which applies modality-specific masking to the visual, shared, and textual subspaces to preserve cross-modal grounding during sequential adaptation. Experiments on linguistic incremental VQA v2 and scene incremental GQA under identical memory constraints show that AIM consistently improves average performance and reduces forgetting, and the VQA v2 compositional results, together with ablation and task-level analyses, further confirm that asymmetric control of plasticity is more effective than treating all trainable parameters uniformly.


\bibliographystyle{ACM-Reference-Format}
\bibliography{references}

@inproceedings{blip,
  title={BLIP: Bootstrapping Language-Image Pre-Training for Unified Vision-Language Understanding and Generation},
  author={Li, Junyang and Li, Dong and Xiong, Chen and Hoi, Steven C. H.},
  booktitle={Proceedings of the International Conference on Machine Learning (ICML)},
  year={2022},
  pages={12888-12900}
}

@InProceedings{pmlr-v202-li23q,
  title = 	 {{BLIP}-2: Bootstrapping Language-Image Pre-training with Frozen Image Encoders and Large Language Models},
  author =       {Li, Junnan and Li, Dongxu and Savarese, Silvio and Hoi, Steven},
  booktitle = 	 {Proceedings of the 40th International Conference on Machine Learning},
  pages = 	 {19730--19742},
  year = 	 {2023},
  editor = 	 {Krause, Andreas and Brunskill, Emma and Cho, Kyunghyun and Engelhardt, Barbara and Sabato, Sivan and Scarlett, Jonathan},
  volume = 	 {202},
  series = 	 {Proceedings of Machine Learning Research},
  month = 	 {23--29 Jul},
  publisher =    {PMLR},
  url = 	 {https://proceedings.mlr.press/v202/li23q.html}
}

@InProceedings{pmlr-v139-radford21a,
  title = 	 {Learning Transferable Visual Models From Natural Language Supervision},
  author =       {Radford, Alec and Kim, Jong Wook and Hallacy, Chris and Ramesh, Aditya and Goh, Gabriel and Agarwal, Sandhini and Sastry, Girish and Askell, Amanda and Mishkin, Pamela and Clark, Jack and Krueger, Gretchen and Sutskever, Ilya},
  booktitle = 	 {Proceedings of the 38th International Conference on Machine Learning},
  pages = 	 {8748--8763},
  year = 	 {2021},
  editor = 	 {Meila, Marina and Zhang, Tong},
  volume = 	 {139},
  series = 	 {Proceedings of Machine Learning Research},
  month = 	 {18--24 Jul},
  publisher =    {PMLR},
  url = 	 {https://proceedings.mlr.press/v139/radford21a.html}
}

@misc{alayrac2022flamingovisuallanguagemodel,
      title={Flamingo: a Visual Language Model for Few-Shot Learning}, 
      author={Jean-Baptiste Alayrac and Jeff Donahue and Pauline Luc and Antoine Miech and Iain Barr and Yana Hasson and Karel Lenc and Arthur Mensch and Katie Millican and Malcolm Reynolds and Roman Ring and Eliza Rutherford and Serkan Cabi and Tengda Han and Zhitao Gong and Sina Samangooei and Marianne Monteiro and Jacob Menick and Sebastian Borgeaud and Andrew Brock and Aida Nematzadeh and Sahand Sharifzadeh and Mikolaj Binkowski and Ricardo Barreira and Oriol Vinyals and Andrew Zisserman and Karen Simonyan},
      year={2022},
      eprint={2204.14198},
      archivePrefix={arXiv},
      primaryClass={cs.CV},
      url={https://arxiv.org/abs/2204.14198}, 
}

@inproceedings{visual_instruction_tuning,
  title={Visual Instruction Tuning},
  author={Liu, H. and Li, C. and Wu, Q. and Lee, Y. J. and Zhang, L. and Wang, W. Y.},
  booktitle={Advances in Neural Information Processing Systems (NeurIPS)},
  volume={36},
  year={2023}
}

@InProceedings{Goyal_2017_CVPR,
author = {Goyal, Yash and Khot, Tejas and Summers-Stay, Douglas and Batra, Dhruv and Parikh, Devi},
title = {Making the v in VQA Matter: Elevating the Role of Image Understanding in Visual Question Answering},
booktitle = {Proceedings of the IEEE Conference on Computer Vision and Pattern Recognition (CVPR)},
month = {July},
year = {2017}
}

@INPROCEEDINGS{8953451,
  author={Hudson, Drew A. and Manning, Christopher D.},
  booktitle={2019 IEEE/CVF Conference on Computer Vision and Pattern Recognition (CVPR)}, 
  title={GQA: A New Dataset for Real-World Visual Reasoning and Compositional Question Answering}, 
  year={2019},
  volume={},
  number={},
  pages={6693-6702},
  doi={10.1109/CVPR.2019.00686}}

@InProceedings{Yu_2019_CVPR,
author = {Yu, Zhou and Yu, Jun and Cui, Yuhao and Tao, Dacheng and Tian, Qi},
title = {Deep Modular Co-Attention Networks for Visual Question Answering},
booktitle = {Proceedings of the IEEE/CVF Conference on Computer Vision and Pattern Recognition (CVPR)},
month = {June},
year = {2019}
}

@inproceedings{ALBEF,
      title={Align before Fuse: Vision and Language Representation Learning with Momentum Distillation}, 
      author={Junnan Li and Ramprasaath R. Selvaraju and Akhilesh Deepak Gotmare and Shafiq Joty and Caiming Xiong and Steven Hoi},
      year={2021},
      booktitle={NeurIPS},
}

@article{Yan_Shou_Ge_Wang_Lin_Cai_Tang_2023, title={Video-Text Pre-training with Learned Regions for Retrieval}, volume={37}, url={https://ojs.aaai.org/index.php/AAAI/article/view/25414}, DOI={10.1609/aaai.v37i3.25414}, abstractNote={Video-Text pre-training aims at learning transferable representations from large-scale video-text pairs via aligning the semantics between visual and textual information. State-of-the-art approaches extract visual features from raw pixels in an end-to-end fashion. However, these methods operate at frame-level directly and thus overlook the spatio-temporal structure of objects in video, which yet has a strong synergy with nouns in textual descriptions. In this work, we propose a simple yet effective module for video-text representation learning, namely RegionLearner, which can take into account the structure of objects during pre-training on large-scale video-text pairs. Given a video, our module (1) first quantizes continuous visual features via clustering patch-features into the same cluster according to content similarity, then (2) generates learnable masks to aggregate fragmentary features into regions with complete semantics, and finally (3) models the spatio-temporal dependencies between different semantic regions. In contrast to using off-the-shelf object detectors, our proposed module does not require explicit supervision and is much more computationally efficient. We pre-train the proposed approach on the public WebVid2M and CC3M datasets. Extensive evaluations on four downstream video-text retrieval benchmarks clearly demonstrate the effectiveness of our RegionLearner.}, number={3}, journal={Proceedings of the AAAI Conference on Artificial Intelligence}, author={Yan, Rui and Shou, Mike Zheng and Ge, Yixiao and Wang, Jinpeng and Lin, Xudong and Cai, Guanyu and Tang, Jinhui}, year={2023}, month={Jun.}, pages={3100-3108} }

@article{Liu2025ContinualLF,
  title={Continual Learning for VLMs: A Survey and Taxonomy Beyond Forgetting},
  author={Yuyang Liu and Qiuhe Hong and Linlan Huang and Alex Gomez-Villa and Dipam Goswami and Xialei Liu and Joost van de Weijer and Yonghong Tian},
  journal={ArXiv},
  year={2025},
  volume={abs/2508.04227},
  url={https://api.semanticscholar.org/CorpusID:280536992}
}

@article{french1999catastrophic,
  title={Catastrophic forgetting in connectionist networks},
  author={French, R. M.},
  journal={Trends in Cognitive Sciences},
  volume={3},
  number={4},
  pages={128-135},
  year={1999}
}

@InProceedings{zhang2023vqacl,
  author    = {Zhang, Xi and Zhang, Feifei and Xu, Changsheng},
  title     = {VQACL: A Novel Visual Question Answering Continual Learning Setting},
  booktitle = {Proceedings of the IEEE/CVF Conference on Computer Vision and Pattern Recognition (CVPR)},
  month     = {June},
  year      = {2023}
}

@article{doi:10.1073/pnas.1611835114,
author = {James Kirkpatrick  and Razvan Pascanu  and Neil Rabinowitz  and Joel Veness  and Guillaume Desjardins  and Andrei A. Rusu  and Kieran Milan  and John Quan  and Tiago Ramalho  and Agnieszka Grabska-Barwinska  and Demis Hassabis  and Claudia Clopath  and Dharshan Kumaran  and Raia Hadsell },
title = {Overcoming catastrophic forgetting in neural networks},
journal = {Proceedings of the National Academy of Sciences},
volume = {114},
number = {13},
pages = {3521-3526},
year = {2017},
doi = {10.1073/pnas.1611835114},
URL = {https://www.pnas.org/doi/abs/10.1073/pnas.1611835114},
eprint = {https://www.pnas.org/doi/pdf/10.1073/pnas.1611835114}}

@ARTICLE{8107520,
  author={Li, Zhizhong and Hoiem, Derek},
  journal={IEEE Transactions on Pattern Analysis and Machine Intelligence}, 
  title={Learning without Forgetting}, 
  year={2018},
  volume={40},
  number={12},
  pages={2935-2947},
  doi={10.1109/TPAMI.2017.2773081}}

@InProceedings{Aljundi_2018_ECCV,
author = {Aljundi, Rahaf and Babiloni, Francesca and Elhoseiny, Mohamed and Rohrbach, Marcus and Tuytelaars, Tinne},
title = {Memory Aware Synapses: Learning what (not) to forget },
booktitle = {Proceedings of the European Conference on Computer Vision (ECCV)},
month = {September},
year = {2018}
}

@InProceedings{pmlr-v80-schwarz18a,
  title = 	 {Progress {\&}amp; Compress: A scalable framework for continual learning},
  author =       {Schwarz, Jonathan and Czarnecki, Wojciech and Luketina, Jelena and Grabska-Barwinska, Agnieszka and Teh, Yee Whye and Pascanu, Razvan and Hadsell, Raia},
  booktitle = 	 {Proceedings of the 35th International Conference on Machine Learning},
  pages = 	 {4528--4537},
  year = 	 {2018},
  editor = 	 {Dy, Jennifer and Krause, Andreas},
  volume = 	 {80},
  series = 	 {Proceedings of Machine Learning Research},
  month = 	 {10--15 Jul},
  publisher =    {PMLR},
  url = 	 {https://proceedings.mlr.press/v80/schwarz18a.html}
}

@article{Robins1995CatastrophicFR,
  title={Catastrophic Forgetting, Rehearsal and Pseudorehearsal},
  author={Anthony V. Robins},
  journal={Connect. Sci.},
  year={1995},
  volume={7},
  pages={123-146},
  url={https://api.semanticscholar.org/CorpusID:22882861}
}

@inproceedings{10.5555/3495724.3497059,
author = {Buzzega, Pietro and Boschini, Matteo and Porrello, Angelo and Abati, Davide and Calderara, Simone},
title = {Dark experience for general continual learning: a strong, simple baseline},
year = {2020},
isbn = {9781713829546},
publisher = {Curran Associates Inc.},
address = {Red Hook, NY, USA},
booktitle = {Proceedings of the 34th International Conference on Neural Information Processing Systems},
articleno = {1335},
numpages = {11},
location = {Vancouver, BC, Canada},
series = {NIPS '20}
}

@InProceedings{Wan_2022_CVPR,
    author    = {Wan, Timmy S. T. and Chen, Jun-Cheng and Wu, Tzer-Yi and Chen, Chu-Song},
    title     = {Continual Learning for Visual Search With Backward Consistent Feature Embedding},
    booktitle = {Proceedings of the IEEE/CVF Conference on Computer Vision and Pattern Recognition (CVPR)},
    month     = {June},
    year      = {2022},
    pages     = {16702-16711}
}

@inproceedings{yang2024quad,
  title={QUAD: Question-only replay for anti-forgetting in visual question answering},
  author={Yang, S. and Chen, L. and Zhang, H. and Xiao, T.},
  booktitle={Proceedings of the IEEE/CVF Conference on Computer Vision and Pattern Recognition (CVPR)},
  pages={22545-22554},
  year={2024}
}

@InProceedings{Huai_2025_CVPR,
    author    = {Huai, Tianyu and Zhou, Jie and Wu, Xingjiao and Chen, Qin and Bai, Qingchun and Zhou, Ze and He, Liang},
    title     = {CL-MoE: Enhancing Multimodal Large Language Model with Dual Momentum Mixture-of-Experts for Continual Visual Question Answering},
    booktitle = {Proceedings of the IEEE/CVF Conference on Computer Vision and Pattern Recognition (CVPR)},
    month     = {June},
    year      = {2025},
    pages     = {19608-19617}
}

@InProceedings{Wang_2020_CVPR,
author = {Wang, Weiyao and Tran, Du and Feiszli, Matt},
title = {What Makes Training Multi-Modal Classification Networks Hard?},
booktitle = {Proceedings of the IEEE/CVF Conference on Computer Vision and Pattern Recognition (CVPR)},
month = {June},
year = {2020}
}

@InProceedings{Peng_2022_CVPR,
    author    = {Peng, Xiaokang and Wei, Yake and Deng, Andong and Wang, Dong and Hu, Di},
    title     = {Balanced Multimodal Learning via On-the-Fly Gradient Modulation},
    booktitle = {Proceedings of the IEEE/CVF Conference on Computer Vision and Pattern Recognition (CVPR)},
    month     = {June},
    year      = {2022},
    pages     = {8238-8247}
}

@inproceedings{10.5555/3600270.3601550,
author = {Liang, Weixin and Zhang, Yuhui and Kwon, Yongchan and Yeung, Serena and Zou, James},
title = {Mind the gap: understanding the modality gap in multi-modal contrastive representation learning},
year = {2022},
isbn = {9781713871088},
publisher = {Curran Associates Inc.},
address = {Red Hook, NY, USA},
booktitle = {Proceedings of the 36th International Conference on Neural Information Processing Systems},
articleno = {1280},
numpages = {14},
location = {New Orleans, LA, USA},
series = {NIPS '22}
}

@InProceedings{Dai_2024_CVPR,
    author    = {Dai, Yusheng and Chen, Hang and Du, Jun and Wang, Ruoyu and Chen, Shihao and Wang, Haotian and Lee, Chin-Hui},
    title     = {A Study of Dropout-Induced Modality Bias on Robustness to Missing Video Frames for Audio-Visual Speech Recognition},
    booktitle = {Proceedings of the IEEE/CVF Conference on Computer Vision and Pattern Recognition (CVPR)},
    month     = {June},
    year      = {2024},
    pages     = {27445-27455}
}

@inproceedings{nguyen2023clove,
  title={CLOVE: A benchmark for continual learning in visual question answering over domain shifts},
  author={Nguyen, T. and Vu, M. N. and Vuong, A. and Nguyen, D. and Vo, T. and Le, N. and Nguyen, A.},
  booktitle={Proceedings of the IEEE/CVF International Conference on Computer Vision (ICCV)},
  pages={1563-1573},
  year={2023}
}

@article{JMLR:v21:17-678,
  author  = {James Martens},
  title   = {New Insights and Perspectives on the Natural Gradient Method},
  journal = {Journal of Machine Learning Research},
  year    = {2020},
  volume  = {21},
  number  = {146},
  pages   = {1--76},
  url     = {http://jmlr.org/papers/v21/17-678.html}
}

@article{wang2024survey,
  author    = {Wang, Liyuan and Zhang, Xingxing and Su, Hang and Zhu, Jun},
  title     = {A Comprehensive Survey of Continual Learning: Theory, Method and Application},
  journal   = {IEEE Transactions on Pattern Analysis and Machine Intelligence},
  volume    = {46},
  number    = {8},
  pages     = {5362--5383},
  year      = {2024}
}

@inproceedings{rebuffi2017icarl,
  author    = {Rebuffi, Sylvestre-Alvise and Kolesnikov, Alexander and Sperl, Georg and Lampert, Christoph H.},
  title     = {{iCaRL}: Incremental Classifier and Representation Learning},
  booktitle = {Proceedings of the IEEE Conference on Computer Vision and Pattern Recognition},
  pages     = {2001--2010},
  year      = {2017}
}

@inproceedings{cha2021co2l,
  author    = {Cha, Hyuntak and Lee, Jaeho and Shin, Jinwoo},
  title     = {{Co2L}: Contrastive Continual Learning},
  booktitle = {Proceedings of the IEEE/CVF International Conference on Computer Vision},
  pages     = {9516--9525},
  year      = {2021}
}

@inproceedings{fini2022self,
  author    = {Fini, Enrico and da Costa, Victor G. Turrisi and Alameda-Pineda, Xavier and Ricci, Elisa and Alahari, Karteek and Mairal, Julien},
  title     = {Self-Supervised Models Are Continual Learners},
  booktitle = {Proceedings of the IEEE/CVF Conference on Computer Vision and Pattern Recognition},
  pages     = {9621--9630},
  year      = {2022}
}

@misc{liu2025vlmsurvey,
  title={Continual Learning for VLMs: A Survey and Taxonomy Beyond Forgetting},
  author={Yuyang Liu and Qiuhe Hong and Linlan Huang and Alexandra Gomez-Villa and Dipam Goswami and Xialei Liu and Joost van de Weijer and Yonghong Tian},
  year={2025},
  note={arXiv preprint}
}

@article{zhou2022coop,
  title={Learning to Prompt for Vision-Language Models},
  author={Zhou, Kaiyang and Yang, Jingkang and Loy, Chen Change and Liu, Ziwei},
  journal={International Journal of Computer Vision},
  volume={130},
  number={9},
  pages={2337--2348},
  year={2022}
}

@article{gao2024clipadapter,
  title={CLIP-Adapter: Better Vision-Language Models with Feature Adapters},
  author={Gao, Peng and Geng, Shijie and Zhang, Renrui and others},
  journal={International Journal of Computer Vision},
  volume={132},
  pages={581--595},
  year={2024},
  doi={10.1007/s11263-023-01891-x}
}

@inproceedings{10.5555/3295222.3295393,
author = {Lopez-Paz, David and Ranzato, Marc'Aurelio},
title = {Gradient episodic memory for continual learning},
year = {2017},
isbn = {9781510860964},
publisher = {Curran Associates Inc.},
address = {Red Hook, NY, USA},
booktitle = {Proceedings of the 31st International Conference on Neural Information Processing Systems},
pages = {6470–6479},
numpages = {10},
location = {Long Beach, California, USA},
series = {NIPS'17}
}

@incollection{zhang2022tipadapter,
  title={Tip-Adapter: Training-Free Adaption of CLIP for Few-Shot Classification},
  author={Zhang, Renrui and Fang, Rongyao and Zhang, Wei and Gao, Peng and Li, Kai and Dai, Jianfei and Qiao, Yu and others},
  booktitle={Computer Vision -- ECCV 2022},
  series={Lecture Notes in Computer Science},
  volume={13695},
  pages={493--510},
  year={2022},
  publisher={Springer},
  doi={10.1007/978-3-031-19833-5_29}
}

@inproceedings{gu2021open,
  title={Open-Vocabulary Object Detection via Vision and Language Knowledge Distillation},
  author={Gu, Xiuye and Lin, Tsung-Yi and Kuo, Weicheng and Cui, Yin},
  booktitle={International Conference on Learning Representations},
  year={2021}
}

@inproceedings{du2022learning,
  title={Learning to Prompt for Open-Vocabulary Object Detection with Vision-Language Model},
  author={Du, Yu and Wei, Fangyun and Zhang, Zihe and Shi, Miaojing and Gao, Yue and Li, Guoqi},
  booktitle={Proceedings of the IEEE/CVF Conference on Computer Vision and Pattern Recognition},
  year={2022}
}

@misc{singh2021flava,
  title={FLAVA: A Foundational Language And Vision Alignment Model},
  author={Amanpreet Singh and Ronghang Hu and Vedanuj Goswami and others},
  year={2021},
  doi={10.48550/arXiv.2112.04482},
  note={arXiv preprint}
}

@incollection{mccloskey1989catastrophic,
  title={Catastrophic interference in connectionist networks: The sequential learning problem},
  author={McCloskey, M. and Cohen, N. J.},
  booktitle={Psychology of Learning and Motivation},
  volume={24},
  pages={109-165},
  year={1989},
  publisher={Academic Press}
}

@misc{buzzega2020darkexperiencegeneralcontinual,
  title={Dark Experience for General Continual Learning: a Strong, Simple Baseline},
  author={Pietro Buzzega and Matteo Boschini and Angelo Porrello and Davide Abati and Simone Calderara},
  year={2020},
  eprint={2004.07211},
  archivePrefix={arXiv},
  primaryClass={stat.ML},
  url={https://arxiv.org/abs/2004.07211}
}

@misc{ren2016fasterrcnnrealtimeobject,
  title={Faster R-CNN: Towards Real-Time Object Detection with Region Proposal Networks},
  author={Shaoqing Ren and Kaiming He and Ross Girshick and Jian Sun},
  year={2016},
  eprint={1506.01497},
  archivePrefix={arXiv},
  primaryClass={cs.CV},
  url={https://arxiv.org/abs/1506.01497}
}

@misc{cho2021unifyingvisionandlanguagetaskstext,
      title={Unifying Vision-and-Language Tasks via Text Generation}, 
      author={Jaemin Cho and Jie Lei and Hao Tan and Mohit Bansal},
      year={2021},
      eprint={2102.02779},
      archivePrefix={arXiv},
      primaryClass={cs.CL},
      url={https://arxiv.org/abs/2102.02779}, 
}

@misc{krishna2016visualgenomeconnectinglanguage,
  title={Visual Genome: Connecting Language and Vision Using Crowdsourced Dense Image Annotations},
  author={Ranjay Krishna and Yuke Zhu and Oliver Groth and Justin Johnson and Kenji Hata and Joshua Kravitz and Stephanie Chen and Yannis Kalantidis and Li-Jia Li and David A. Shamma and Michael S. Bernstein and Fei-Fei Li},
  year={2016},
  eprint={1602.07332},
  archivePrefix={arXiv},
  primaryClass={cs.CV},
  url={https://arxiv.org/abs/1602.07332}
}

@misc{marouf2025askrememberquestionsonlyreplay,
  title={Ask and Remember: A Questions-Only Replay Strategy for Continual Visual Question Answering},
  author={Imad Eddine Marouf and Enzo Tartaglione and Stephane Lathuiliere and Joost van de Weijer},
  year={2025},
  eprint={2502.04469},
  archivePrefix={arXiv},
  primaryClass={cs.CV},
  url={https://arxiv.org/abs/2502.04469}
}

@inbook{Chaudhry_2018,
  title={Riemannian Walk for Incremental Learning: Understanding Forgetting and Intransigence},
  ISBN={9783030012526},
  ISSN={1611-3349},
  url={http://dx.doi.org/10.1007/978-3-030-01252-6_33},
  DOI={10.1007/978-3-030-01252-6_33},
  booktitle={Computer Vision – ECCV 2018},
  publisher={Springer International Publishing},
  author={Chaudhry, Arslan and Dokania, Puneet K. and Ajanthan, Thalaiyasingam and Torr, Philip H. S.},
  year={2018},
  pages={556–572}
}

@misc{loshchilov2019decoupledweightdecayregularization,
      title={Decoupled Weight Decay Regularization}, 
      author={Ilya Loshchilov and Frank Hutter},
      year={2019},
      eprint={1711.05101},
      archivePrefix={arXiv},
      primaryClass={cs.LG},
      url={https://arxiv.org/abs/1711.05101}, 
}

@inproceedings{48153,title	= {BERT Rediscovers the Classical NLP Pipeline},author	= {Ian Tenney and Dipanjan Das and Ellie Pavlick},year	= {2019},URL	= {https://arxiv.org/abs/1905.05950},booktitle	= {Association for Computational Linguistics}}

@inproceedings{Jawahar2019WhatDB,
  title={What Does BERT Learn about the Structure of Language?},
  author={Ganesh Jawahar and Beno{\^i}t Sagot and Djam{\'e} Seddah},
  booktitle={Annual Meeting of the Association for Computational Linguistics},
  year={2019},
  url={https://api.semanticscholar.org/CorpusID:195477534}
}

@inproceedings{li2020what,
  title={What Does BERT with Vision Look At?},
  author={Li, Liunian Harold and Yatskar, Mark and Yin, Da and Hsieh, Cho-Jui and Chang, Kai-Wei},
  booktitle={Proceedings of the 58th Annual Meeting of the Association for Computational Linguistics (ACL)},
  pages={5265--5275},
  year={2020}
}

@inproceedings{Cao2020BehindTS,
  title={Behind the Scene: Revealing the Secrets of Pre-trained Vision-and-Language Models},
  author={Jize Cao and Zhe Gan and Yu Cheng and Licheng Yu and Yen-Chun Chen and Jingjing Liu},
  booktitle={European Conference on Computer Vision},
  year={2020},
  url={https://api.semanticscholar.org/CorpusID:218665405}
}

@misc{li2023blip2bootstrappinglanguageimagepretraining,
      title={BLIP-2: Bootstrapping Language-Image Pre-training with Frozen Image Encoders and Large Language Models}, 
      author={Junnan Li and Dongxu Li and Silvio Savarese and Steven Hoi},
      year={2023},
      eprint={2301.12597},
      archivePrefix={arXiv},
      primaryClass={cs.CV},
      url={https://arxiv.org/abs/2301.12597}, 
}

@misc{liu2023visualinstructiontuning,
      title={Visual Instruction Tuning}, 
      author={Haotian Liu and Chunyuan Li and Qingyang Wu and Yong Jae Lee},
      year={2023},
      eprint={2304.08485},
      archivePrefix={arXiv},
      primaryClass={cs.CV},
      url={https://arxiv.org/abs/2304.08485}, 
}

@InProceedings{Agrawal_2018_CVPR,
author = {Agrawal, Aishwarya and Batra, Dhruv and Parikh, Devi and Kembhavi, Aniruddha},
title = {Don't Just Assume; Look and Answer: Overcoming Priors for Visual Question Answering},
booktitle = {Proceedings of the IEEE Conference on Computer Vision and Pattern Recognition (CVPR)},
month = {June},
year = {2018}
}

@inproceedings{inproceedings,
author = {Wolf, Thomas and Debut, Lysandre and Sanh, Victor and Chaumond, Julien and Delangue, Clement and Moi, Anthony and Cistac, Pierric and Rault, Tim and Louf, Remi and Funtowicz, Morgan and Davison, Joe and Shleifer, Sam and Platen, Patrick and Ma, Clara and Jernite, Yacine and Plu, Julien and Xu, Canwen and Scao, Teven and Gugger, Sylvain and Rush, Alexander},
year = {2020},
month = {01},
pages = {38-45},
title = {Transformers: State-of-the-Art Natural Language Processing},
doi = {10.18653/v1/2020.emnlp-demos.6}
}



\clearpage
\appendix

\section*{Supplementary Material Overview}

\noindent This supplement details our Asymmetric Information Masking (AIM) for visual question answering continual learning, which mitigates catastrophic forgetting in sequential task adaptation.Our key insight is to asymmetrically regulate parameter updates across modality-specific subspaces, thereby preserving learned vision-language knowledge while enabling adaptation to new tasks.We provide full details on dataset construction, implementation, and extended experiments in the sections that follow. We first describe the dataset construction and evaluation protocols, then give fuller descriptions of the baselines and implementation settings. After that, we present the extension of AIM to a large multimodal model, followed by additional ablations, parameter analyses, efficiency results, and qualitative case studies.

\noindent The supplementary is organized as follows:
\begin{itemize}
    \item In Sec.~A, we describe the detailed dataset setup and evaluation protocols for VQA v2 and GQA. We also summarize the task statistics and the construction of the continual learning splits used in our experiments.
    
    \item Sec.~B reviews the baseline in our comparisons. We include EWC, LwF, MAS, ER, DER, VQACL and QUAD as baselines, all implemented under the same training protocol.
    
    \item Sec.~C details the implementation of AIM. We detail the architecture and training procedure, including hyperparameter settings and memory configuration.
    
    \item Sec.~D extends AIM to the LLaVA architecture. This section explains the parameter partition in LLaVA, the training setup, and the corresponding results on VQA v2.
    
    \item Sec.~E collects additional ablation and sensitivity studies. It examines stability and plasticity, compositional generalization, and robustness under different task orders.
    
    \item Sec.~F takes a closer look at the asymmetry dynamics of AIM. We analyze parameter shifts, Fisher-based sensitivity, and layer-wise behavior across the continual learning process.
    
    \item Sec.~G quantifies the extra cost introduced by AIM. We measure the overhead in optimization volume, wall-clock time, and GPU memory usage.
    
    \item Sec.~H turns to qualitative case studies. By comparing AIM with the Vanilla baseline, it shows how the two models differ in sample-level grounding after sequential task shifts.
\end{itemize}

\section{Detailed Dataset Setup and Protocols}
\label{supp:datasets}

In the main paper, we evaluated our method on two multimodal benchmarks that stress continual learning from different angles. VQA v2 is used to study linguistic-incremental learning, where the task sequence is organized according to question types and the model must preserve previously acquired reasoning skills while adapting to new ones. GQA is used to study scene-incremental learning, where the task sequence follows large scene-level distribution shifts and mainly tests the robustness of visual grounding under changing environments.

\paragraph{VQA v2 (Linguistic-Incremental).}
We use the VQA v2 benchmark~\cite{Goyal_2017_CVPR} and follow the VQACL protocol~\cite{zhang2023vqacl} to construct the continual learning sequence. The benchmark is partitioned into 10 mutually exclusive tasks according to question types, such as \textit{Counting}, \textit{Color}, and \textit{Location}. Each task emphasizes a different reasoning skill, and the full sequence evaluates whether the model can continuously acquire new question-answering capabilities without forgetting earlier ones. For evaluation, we report both the standard test performance and the Novel Composition (COMP) performance. In the COMP setting, we adopt the 5-fold object-independent cross-validation protocol introduced in VQACL~\cite{zhang2023vqacl}, where one object group is held out from training for each fold and only appears at test time in combination with the corresponding reasoning skill. This protocol ensures that evaluation samples contain unseen skill-concept compositions rather than repetitions of training patterns.

\begin{table*}[t]
\centering
\caption{Linguistic-driven task statistics of VQA v2 in the VQACL setting. Stan. Test denotes the standard test set.}
\label{tab:vqav2_task_stats}
\resizebox{\textwidth}{!}{
\begin{tabular}{lcccccc p{8.8cm}}
\toprule
\textbf{Task} & \textbf{Train} & \textbf{Train Unique} & \textbf{Val} & \textbf{Val Unique} & \textbf{Stan. Test} & \textbf{Test Unique} & \textbf{Examples} \\
\midrule
Recognition & 146,425 & 52,730 & 6,213 & 4,780 & 6,265 & 4,831 & What is written on top of the door? \\
Location & 14,921 & 3,677 & 710 & 442 & 702 & 454 & Where are the birds standing? \\
Judge & 178,990 & 68,390 & 7,955 & 6,427 & 7,970 & 6,468 & Is the animal using the keyboard? \\
Commonsense & 28,312 & 14,466 & 1,269 & 1,157 & 1,227 & 1,128 & Does the female look angry? \\
Count & 68,564 & 18,678 & 2,930 & 2,050 & 2,905 & 2,060 & How many discrete orange patches are on the cat? \\
Action & 35,185 & 10,759 & 1,572 & 1,090 & 1,448 & 1,029 & Is he or she skating on the sidewalk? \\
Color & 57,303 & 10,902 & 2,590 & 1,341 & 2,451 & 1,272 & What color are the awning? \\
Type & 26,563 & 8,534 & 1,260 & 883 & 1,211 & 888 & What type of plant life is featured in the picture? \\
Subcategory & 35,880 & 13,886 & 1,681 & 1,245 & 1,604 & 1,180 & Did the people just depart the plane? \\
Causal & 6,376 & 3,788 & 250 & 238 & 217 & 208 & Why is she reaching up? \\
\bottomrule
\end{tabular}
}
\end{table*}

\begin{table*}[t]
\centering
\caption{Detailed information about the five object groups in VQA v2.}
\label{tab:vqav2_object_groups}
\resizebox{\textwidth}{!}{
\begin{tabular}{lp{15.5cm}}
\toprule
\textbf{Task} & \textbf{Objects} \\
\midrule
Group 1 & hot dog, fork, orange, snowboard, potted plant, person, toilet, laptop, surfboard, bench, bus, dog, knife, pizza, handbag, bicycle \\
Group 2 & horse, cell phone, elephant, boat, zebra, apple, stop sign, microwave, spoon, cup, skateboard, tie, umbrella, sandwich, bear \\
Group 3 & donut, truck, frisbee, giraffe, dining table, motorcycle, parking meter, car, oven, airplane, bed, sheep, baseball bat \\
Group 4 & skis, baseball glove, tennis racket, tv, traffic light, kite, cake, keyboard, bottle, remote, bird, carrot \\
Group 5 & suitcase, couch, broccoli, cow, fire hydrant, chair, mouse, cat, banana, wine glass, backpack, bowl, sports ball, train \\
\bottomrule
\end{tabular}
}
\end{table*}

Table~\ref{tab:vqav2_task_stats} summarizes the task-level statistics and representative question examples.

\paragraph{GQA (Scene-Incremental).}
We use GQA~\cite{8953451} as the benchmark for scene-incremental continual learning and follow the CLOVE benchmark~\cite{nguyen2023clove} to simulate stronger visual domain shifts. Specifically, the dataset is partitioned into 6 scene-based tasks: \textit{Shop \& Dining}, \textit{Workplace}, \textit{Home}, \textit{Transportation}, \textit{Sports}, and \textit{Outdoors}. These tasks are learned sequentially, so the model must adapt to changing scene distributions while retaining the visual grounding knowledge learned from earlier environments. Unlike VQA v2, where the main source of variation comes from reasoning skills, GQA mainly stresses scene-level appearance changes and cross-domain transfer of visual-linguistic alignment.

\begin{table*}[t]
\centering
\caption{Scene-driven task statistics of GQA in the CLOVE-style scene-incremental setting.}
\label{tab:gqa_task_stats}
\resizebox{\textwidth}{!}{
\begin{tabular}{lcccc p{8.8cm}}
\toprule
\textbf{Task} & \textbf{Train} & \textbf{Train Unique} & \textbf{Val} & \textbf{Val Unique} & \textbf{Examples} \\
\midrule
Shop \& Dining & 20,000 & 16,004 & 3,000 & 2,574 & What is sitting on the cooking utensil to the left of the bottle? \\
Workplace & 20,000 & 14,627 & 3,000 & 2,543 & Is the happy man on the bench? \\
Home & 20,000 & 13,719 & 3,000 & 2,472 & What cooking utensil is on the white stove? \\
Transportation & 20,000 & 12,747 & 3,000 & 2,322 & What do you think is the fence on? \\
Sports & 20,000 & 10,920 & 3,000 & 2,052 & What color are the shoes, red or blue? \\
Outdoors & 20,000 & 11,257 & 3,000 & 2,113 & What color does the jacket have? \\
\bottomrule
\end{tabular}
}
\end{table*}

Table~\ref{tab:gqa_task_stats} reports the corresponding statistics for the six scene-based tasks.

\section{Detailed Descriptions of Baseline Methods}
\label{supp:baselines}

We compare AIM with three groups of continual learning baselines. The first group includes regularization-based methods: EWC~\cite{doi:10.1073/pnas.1611835114}, LwF~\cite{8107520}, and MAS~\cite{Aljundi_2018_ECCV}, which mitigate forgetting by constraining parameter updates. The second group includes replay-based methods: ER~\cite{Robins1995CatastrophicFR} and DER~\cite{10.5555/3495724.3497059}, which retain past knowledge through memory rehearsal. The third group includes continual-VQA-specific methods: VQACL~\cite{zhang2023vqacl} and QUAD~\cite{yang2024quad}, designed for sequential multimodal question answering. For a fair comparison, all methods are implemented under the same backbone and continual learning protocol as described in the main paper. We also report \textbf{Joint Training} as an upper bound and \textbf{Vanilla Fine-tuning} as a lower bound.

\paragraph{EWC~\cite{doi:10.1073/pnas.1611835114}.}
EWC estimates the importance of each parameter with the diagonal Fisher Information Matrix and penalizes changes on important parameters when learning new tasks. In this way, it tries to preserve knowledge that is useful for previous tasks. However, the regularization is applied globally and does not distinguish different modality branches in a multimodal model.

\paragraph{LwF~\cite{8107520}.}
LwF uses the previous model as a teacher and distills its outputs while training on the current task. Since it does not store old samples, it is memory-free and easy to apply in sequential training. Still, the supervision is only imposed on current-task inputs, which limits its ability to revisit past multimodal distributions.

\paragraph{MAS~\cite{Aljundi_2018_ECCV}.}
MAS regularizes parameters that strongly affect the model output by measuring output sensitivity with respect to parameter perturbations. Similar to EWC, it protects important parameters through a global penalty during sequential updates. This allows MAS to estimate parameter importance without relying on label-dependent Fisher computation.

\paragraph{ER~\cite{Robins1995CatastrophicFR}.}
ER stores samples from previous tasks in a fixed-size memory buffer and replays them together with current-task data during training. By revisiting past examples, it directly reduces forgetting and remains a strong rehearsal baseline. Due to its simplicity and stable performance, ER is one of the most common replay-based references in continual learning.

\paragraph{DER~\cite{10.5555/3495724.3497059}.}
DER extends standard replay by additionally matching historical model outputs on stored samples. This extra distillation signal helps preserve past decision boundaries and usually gives stronger retention than plain ER. In practice, DER combines sample replay with output-level regularization during sequential updates.

\paragraph{VQACL~\cite{zhang2023vqacl}.}
VQACL is a continual visual question answering method tailored to the VQACL setting. It uses prototype-based replay to preserve task-specific knowledge and transferable task-invariant representations across tasks. This makes it a strong baseline for continual VQA under the same evaluation protocol.

\paragraph{QUAD~\cite{yang2024quad}.}
QUAD is a recent continual VQA method that replaces full multimodal replay with question-only rehearsal and attention distillation. It stores historical questions instead of complete image-question-answer triplets, reducing memory usage and avoiding retaining past images. It also regularizes sequential learning by preserving output and attention consistency across tasks.

\section{Implementation Details}
\label{supp:implementation}

This section provides the implementation details of our Asymmetric Information Masking (AIM) framework for reproducibility, covering environmental setups, hyperparameters, and the complete training procedure.

\paragraph{Architecture and Environment.}
Our model is implemented with PyTorch (v1.13.1) and the HuggingFace Transformers library (v4.2.1)~\cite{inproceedings}. Following standard VQA continual learning protocols~\cite{zhang2023vqacl, yang2024quad}, we initialize the \textbf{VL-T5} architecture with pre-trained \texttt{t5-base} weights. All experiments are conducted on a single NVIDIA A800 GPU.

\paragraph{Training Hyperparameters.}
Across all sequential tasks, the model is optimized using AdamW with a weight decay of 0.01. The base learning rate is set to $1\times 10^{-4}$ with a linear learning rate scheduler and a warmup ratio of 0.05. Each task is trained for 3 epochs. The effective batch size on a single GPU is 80. To prevent exploding gradients, the gradient clipping norm is bounded at 5.0. During inference, we use beam search with a beam size of 5.

\paragraph{Continual Learning and AIM Settings.}
At the end of each task, we estimate the empirical Fisher Information Matrix to evaluate parameter sensitivity. The Fisher values are computed using a randomly sampled subset of the current task's training data for efficiency. The modality-specific protection masks $M_{i}^{(k)}$ are defined by pruning ratios $\rho_{vis}, \rho_{text}$, and $\rho_{shared}$. Thresholding is applied independently within each respective subspace ($\Theta_{vis}$, $\Theta_{text}$, and $\Theta_{shared}$) to decouple the plasticity constraints. The optimal configuration is set to $\rho_{vis}=0.3$, $\rho_{text}=0.5$, and $\rho_{shared}=0.1$, with the orthogonal projection coefficient $\lambda_{orth}=0.1$. The episodic memory buffer $\mathcal{M}$ maintains a maximum capacity of 5,000 for VQA v2 and 500 for GQA. The buffer is managed via random sampling to update and store historical data.

\paragraph{Algorithm Overview.}
The complete training procedure of AIM, including gradient modulation and modality-specific mask generation, is summarized in Algorithm~\ref{alg:aims}.

\begin{algorithm}[t]
\caption{Asymmetric Information Masking (AIM)}
\label{alg:aims}
\begin{algorithmic}[1]
\REQUIRE Tasks $\{\mathcal{T}_k\}_{k=1}^N$, Params $\Theta$, Ratios $\{\rho_m\}$, Buffer $\mathcal{M}$, LR $\eta$
\ENSURE Learned parameters $\Theta$

\FOR{each task $k \in \{1, \dots, N\}$}
    \STATE \textbf{if} $k > 1$ \textbf{then} Populate $\mathcal{M}$ with samples from $T_{1 \dots k-1}$
    
    \STATE \textit{// 1. Optimization with Gradient Modulation}
    \FOR{each task batch $\mathcal{B}_{task} \in \mathcal{T}_k$}
        \STATE $\mathcal{L} \leftarrow \mathcal{L}_{task} + \mathbb{I}(k>1)\mathcal{L}_{mem}$ \quad \textit{// where $\mathcal{B}_{mem} \sim \mathcal{M}$} 
        \STATE $g_i \leftarrow \nabla_{\theta_i} \mathcal{L}$; \quad $g'_i \leftarrow \mathbb{I}(k=1)g_i + \mathbb{I}(k>1)(g_i \odot M_i)$
        \STATE Update: $\theta_i \leftarrow \theta_i - \eta g'_i, \quad \forall \theta_i \in \Theta$ 
    \ENDFOR
    
    \STATE \textit{// 2. Sensitivity Estimation \& Aggregation}
    \STATE Estimate $F_{k, i}$ using $\mathcal{D}_{sub}$; \quad $F_{i}^{(k)} \leftarrow \max(F_{i}^{(k-1)}, F_{k, i})$ 
    
    \STATE \textit{// 3. Mask Generation via Subspace Quantiles}
    \FOR{each modality subspace $m \in \{vis, shared, text\}$}
        \STATE $\tau_m \leftarrow \text{Quantile}(\mathbf{F}_m, 1 - \rho_m)$ 
        \STATE $M_i \leftarrow \mathbb{I}(F_{i}^{(k)} < \tau_m), \quad \forall \theta_i \in \Theta_m$ 
    \ENDFOR
\ENDFOR
\RETURN $\Theta$
\end{algorithmic}
\end{algorithm}

\section{Generalization to Large Multimodal Models}
\label{supp:llava}
In the main paper, we evaluated AIM using VL-T5 to conduct extensive ablation studies. To verify that the asymmetric plasticity design applies to broader multimodal architectures, we extend the framework to Large Multimodal Models (MLLMs), specifically \textbf{LLaVA}~\cite{liu2023visualinstructiontuning}.

\paragraph{Implementation Details for LLaVA.}
We build upon the LLaVA-1.5 architecture, which utilizes a frozen CLIP visual encoder (ViT-L/14-336px), a trainable two-layer MLP visual projector (\texttt{mlp2x\_gelu}), and the vicuna-7b-v1.5 Large Language Model (LLM) backbone. To adapt the 7B parameter space efficiently during continual learning, we apply Low-Rank Adaptation (LoRA) to all linear layers within the LLM backbone, configuring the rank $r=64$ and $\alpha=128$. The models are trained using the AdamW optimizer with a cosine learning rate scheduler, setting the base learning rate to $2 \times 10^{-4}$ for LoRA modules and $2 \times 10^{-5}$ for the visual projector. All tasks are trained for a single epoch utilizing \texttt{bfloat16} precision.

\paragraph{Asymmetric Partitioning in LLaVA-7B.} 
Unlike VL-T5, treating the entire 7B LLM as a single cognitive subspace is suboptimal due to its depth. Based on the 32-layer Transformer architecture of Vicuna-7B, we map the AIM parameter partitioning as follows:
\begin{itemize}
    \item \textbf{Visual Subspace ($\Theta_{vis}$):} Comprises the parameters of the \texttt{mlp2x\_gelu} visual projector, acting as the adaptable modality aligner.
    \item \textbf{Shared Subspace ($\Theta_{shared}$):} Comprises the LoRA adapters in the first 16 layers (Layers 0--15) of the LLM. These early layers primarily fuse the projected visual tokens with linguistic prompt tokens.
    \item \textbf{Cognitive Reasoning Subspace ($\Theta_{text}$):} Includes the LoRA adapters in the remaining 16 layers (Layers 16--31), the language modeling head (\texttt{lm\_head}), and token embeddings. These deeper layers act as the core reasoning engine and require regularization to mitigate catastrophic forgetting.
\end{itemize}

\paragraph{Detailed Results on VQA v2.}
We benchmark AIM against standard and recent continual learning baselines on the LLaVA-7B architecture across all 10 distinct task types in VQA v2. The baseline results, including the recent \textbf{CL-MoE}~\cite{Huai_2025_CVPR}, are referenced from recent literature under the same memory budget ($M=5000$). 

As detailed in Table~\ref{tab:supp_llava_detailed}, sequentially fine-tuning LLaVA (Vanilla FT) leads to substantial forgetting across most reasoning skills. While CL-MoE~\cite{Huai_2025_CVPR} achieves the highest average performance by leveraging a dynamic Mixture-of-Experts architecture, our AIM framework yields an AP of \textbf{48.15\%} without requiring specialized modular experts or dynamic routing. Furthermore, AIM outperforms the VQACL baseline in 7 out of 10 tasks. This indicates that the proposed asymmetric parameter isolation mechanism remains effective when scaled to modern MLLMs.

\begin{table*}[t]

  \centering

  \caption{Detailed continual learning performance (\%) on VQA v2 using the LLaVA-7B architecture. Baseline results are referenced from \cite{Huai_2025_CVPR} under the same $M=5000$ memory budget setting. Best results are highlighted in \textbf{bold}, and second-best results are \underline{underlined}.}

  \label{tab:supp_llava_detailed}

  \resizebox{\textwidth}{!}{

  \begin{tabular}{l cccccccccc cc}

    \toprule

    \textbf{Method} & \textbf{Rec.} & \textbf{Loc.} & \textbf{Jud.} & \textbf{Com.} & \textbf{Cou.} & \textbf{Act.} & \textbf{Col.} & \textbf{Typ.} & \textbf{Sub.} & \textbf{Cau.} & \textbf{AP ($\uparrow$)} & \textbf{AF ($\downarrow$)} \\

    \midrule

    Vanilla & 19.25 & 14.81 & 54.59 & 56.97 & 24.23 & 46.20 & 27.58 & 26.09 & 36.47 & 18.89 & 32.51 & 20.69 \\

    ER \cite{Robins1995CatastrophicFR} & 29.31 & 25.74 & 63.46 & \underline{65.78} & 31.92 & 58.39 & \textbf{45.17} & 34.55 & 46.24 & 18.96 & 41.95 & 10.20 \\

    VQACL \cite{zhang2023vqacl} & 34.14 & 32.19 & 66.15 & 63.00 & 33.01 & 60.91 & 34.64 & 38.48 & 47.94 & \textbf{24.42} & 43.49 & \underline{9.10} \\

    CL-MoE \cite{Huai_2025_CVPR} & \textbf{46.50} & \textbf{37.18} & \textbf{75.22} & \textbf{71.39} & \textbf{40.90} & \textbf{69.54} & \underline{43.66} & \textbf{52.68} & \textbf{55.55} & 20.74 & \textbf{51.34} & \textbf{-0.02} \\

    \midrule

    \textbf{AIM (Ours)} & \underline{39.17} & \underline{32.62} & \underline{70.75} & 65.04 & \underline{39.38} & \underline{65.68} & 39.86 & \underline{48.22} & \underline{50.69} & \underline{23.04} & \underline{47.45} & 12.15 \\

    \bottomrule

  \end{tabular}

  }

\end{table*}

\section{Extended Ablation and Sensitivity Studies}

\subsection{Extended analysis of Stability and Plasticity}
\label{supp:heatmaps}

\begin{figure*}[t]
  \centering
  \includegraphics[width=\textwidth]{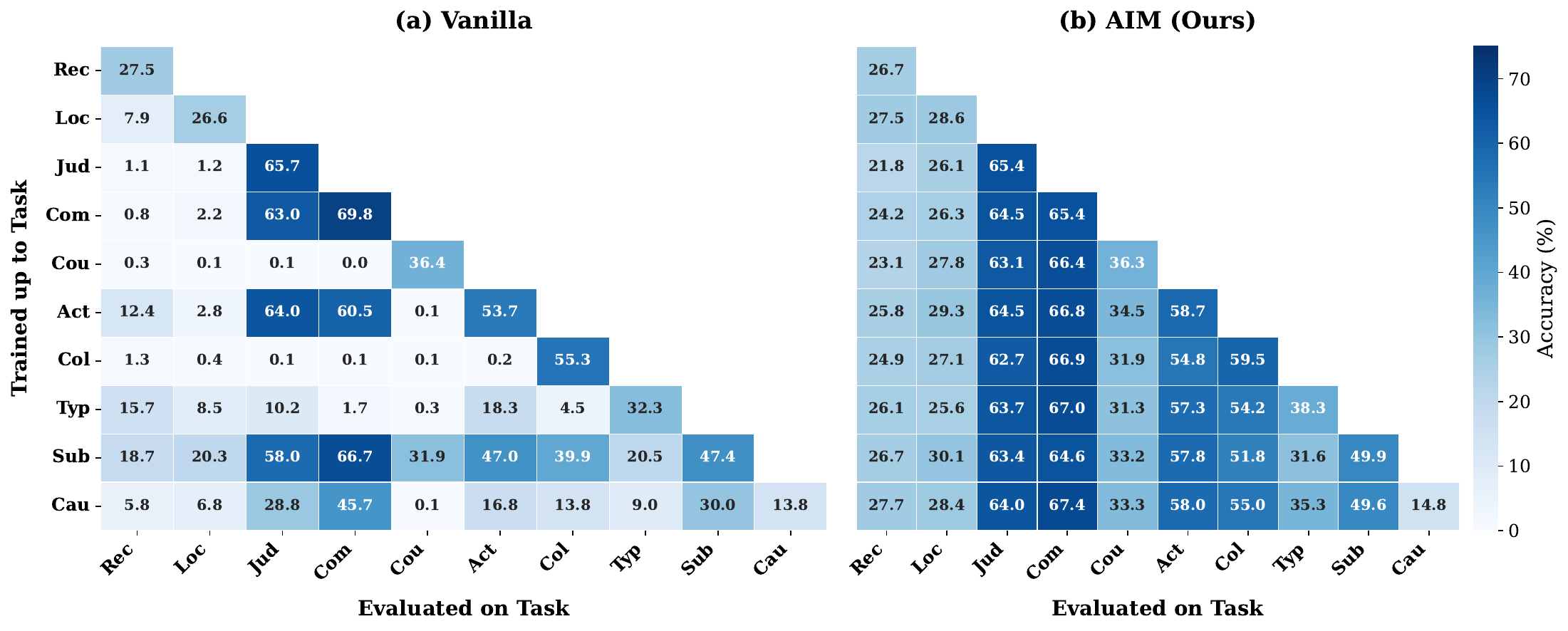}
  \caption{Comprehensive task-by-task evaluation matrices on the VQA v2 benchmark. (a) The \textit{Vanilla} model (based on our highly-optimized reproduction) exhibits severe catastrophic forgetting, as indicated by the fading colors in the vertical columns. (b) \textit{AIM (Ours)} consistently retains performance on previously learned reasoning skills across the extended 10-task sequence.}
  \label{figE:supp_heatmap_vqav2}
\end{figure*}

\textbf{Note on Baseline Reproduction:} Table 1 of the main paper cites the official \textit{Vanilla} results from VQACL~\cite{zhang2023vqacl} for fair literature comparison. Since the original task-by-task evaluation matrices are publicly unavailable, Figure 3 of the main paper and Figure~\ref{figE:supp_heatmap_vqav2} visualize our locally reproduced baseline. Given space constraints in the main text, we present the comprehensive $10 \times 10$ evaluation matrices for the VQA v2 benchmark in Fig.~\ref{figE:supp_heatmap_vqav2}. Due to hardware and random seed variations, our reproduction yields slightly higher overall performance, providing a robust and strictly controlled basis for the dynamic analysis.

As illustrated in Fig.~\ref{figE:supp_heatmap_vqav2}(a), sequentially fine-tuning the \textit{Vanilla} model results in substantial catastrophic forgetting across the 10-task sequence. The performance on early tasks degrades rapidly as training progresses. For instance, the accuracy on the \textit{Recognition} and \textit{Counting} tasks drops drastically (to 5.81\% and a near-zero 0.12\%, respectively) by the end of the continual learning process. 

Conversely, Fig.~\ref{figE:supp_heatmap_vqav2}(b) demonstrates that our AIM framework effectively mitigates this degradation. The lower-triangular elements remain consistent throughout the sequence. Notably, complex reasoning skills acquired early in the training, such as \textit{Judge} and \textit{Commonsense}, retain high accuracy (maintaining over 64\% and 67\%, respectively) even after the model adapts to nine subsequent tasks. This sustained performance across a diverse set of visual-linguistic tasks verifies the robustness of the proposed asymmetric parameter isolation mechanism in extended continual learning scenarios.

\subsection{Detailed Ablation on Compositional Generalization}
\label{supp:ablation_comp}

\paragraph{Evaluation Protocol.} Due to the substantial computational overhead of the full 5-fold cross-validation required for the compositional (COMP) setting, our ablation studies are conducted exclusively on the Group-1 (G1) split. As demonstrated in our main results, the performance on the G1 split provides a sufficient and consistent indicator for evaluating the relative effectiveness of different architectural designs within our framework.

\paragraph{Necessity of Asymmetric Masking.} We compare our optimal asymmetric setting ($\rho_{vis}=0.3, \rho_{text}=0.5$) against two baselines: \textit{Uniform Masking} ($\rho_{vis}=\rho_{text}=0.3$) and \textit{Swapped Masking} ($\rho_{vis}=0.5, \rho_{text}=0.3$). As shown in Table~\ref{tab:ablation_g1_comp}, \textbf{AIM (Ours)} consistently achieves the highest AP across both Novel and Seen compositions. The performance degradation in the Uniform and Swapped variants validates our core hypothesis: visual and linguistic modalities possess different capacities and redundancies. Imposing suboptimal or uniform regularization either restricts the plasticity of the visual bottleneck or under-regularizes the text decoder, thereby impairing the model's compositional generalization.

\paragraph{Impact of Episodic Memory.} To isolate the intrinsic contribution of the masking mechanism, we evaluate AIM without the episodic memory buffer (\textit{w/o Memory, $M=0$}). The removal of experience replay results in a severe performance drop (from 41.92\% to 28.38\% in Novel AP) and a surge in forgetting (AF rises to 17.08\% on Novel compositions). This demonstrates that while asymmetric masking effectively isolates task-specific parameters, episodic memory remains an indispensable component for maintaining performance and preventing catastrophic forgetting of previously aligned visual-linguistic concepts.

\paragraph{Fisher Aggregation Strategy.} We evaluate the \textit{Max Aggregation} operator used in AIM against a \textit{Sum Aggregation} alternative. Interestingly, while \textit{Sum Aggregation} yields a strictly lower AF (1.77\% vs. 3.12\% on Novel compositions), \textit{Max Aggregation} achieves the highest overall AP (41.92\% vs. 41.83\%). This indicates a plasticity-stability trade-off: Max Aggregation better preserves the peak parameter saliency required for specific tasks, which facilitates superior adaptation to novel compositional concepts (higher AP), albeit at the cost of a marginal increase in forgetting.

\begin{table}[t]
\centering
\caption{Detailed ablation study on the Compositional (COMP) setting evaluated on the G1 split. We report AP and AF across \textbf{Novel} and \textbf{Seen} compositions. \textit{Uniform Masking} applies a global ratio $\rho=0.3$. \textit{AIM (Ours)} utilizes the optimal asymmetric ratios ($\rho_{vis}=0.3, \rho_{text}=0.5$) with Max Aggregation.}
\label{tab:ablation_g1_comp}
\resizebox{0.95\columnwidth}{!}{
\begin{tabular}{l | cc | cc}
\toprule
\multirow{2}{*}{\textbf{Method}} & \multicolumn{2}{c|}{\textbf{Novel}} & \multicolumn{2}{c}{\textbf{Seen}} \\
\cmidrule(lr){2-3} \cmidrule(lr){4-5} 
 & \textbf{AP $\uparrow$} & \textbf{AF $\downarrow$} & \textbf{AP $\uparrow$} & \textbf{AF $\downarrow$} \\
\midrule
Uniform Masking        & 39.78 & 2.65 & 42.16 & \underline{1.57} \\
Swapped Masking        & 41.14 & \underline{1.89} & 42.87 & 2.08 \\
AIM w/o Memory         & 28.38 & 17.08 & 27.77 & 18.55 \\
AIM (Sum Agg.)         & 41.83 & \textbf{1.77} & 43.56 & \textbf{1.83} \\
\midrule
\textbf{AIM (Ours)}    & \textbf{41.92} & 3.12 & \textbf{43.66} & 2.09 \\
\bottomrule
\end{tabular}
}
\end{table}

\subsection{Robustness to Task Ordering}
\label{supp:task_ordering}

In continual learning, task ordering can significantly impact a model's ability to generalize to unseen compositions. To assess the robustness of the proposed \textit{Asymmetric Information Masking} (AIM) framework, we evaluate its performance under three distinct task permutations on the VQA v2 benchmark. All metrics, including AP and AF, are reported under the \textbf{Compositional (COMP)} setting. Specifically, the \textbf{Default Order} follows the standard sequence: \{q\_recognition, q\_location, q\_judge, q\_commonsense, q\_count, q \_action, q\_color, q\_type, q\_subcategory, q\_causal\}. The \textbf{Reverse Order} strictly inverts this sequence: \{q\_causal, q\_subcategory, q\_type, q\_color, q\_action, q\_count, q\_commonsense, q\_judge, q\_location, q\_recognition\}. Finally, the \textbf{Random Order} evaluates a shuffled curriculum: \{q\_type, q\_commonsense, q\_judge, q\_subcategory, q\_action, q\_color, q\_causal, q\_count, q\_recognition, q\_location\}.

\paragraph{Quantitative Analysis.} 
Table~\ref{tab:task_order_results} summarizes the results. The \textit{Vanilla} baseline exhibits severe degradation and high variance in compositional reasoning across different sequences, achieving a low average AP of 6.57\% with a standard deviation of $\sigma = 4.69$. This indicates that without explicit parameter-level constraints, the model's compositional generalization remains highly sensitive to the training curriculum.

In contrast, \textbf{AIM} demonstrates superior stability and performance, yielding an average AP of 41.05\% with a minimal standard deviation of $\sigma = 0.96$. Furthermore, AIM dramatically reduces Average Forgetting (AF) to 2.20\% ($\sigma = 1.41$) compared to the Vanilla baseline's 31.43\% ($\sigma = 3.71$). These results indicate that the asymmetric masking mechanism effectively mitigates task-specific interference, ensuring that acquired reasoning capabilities remain remarkably stable across varying training trajectories.

\begin{table}[t]
\centering
\caption{Robustness analysis of task ordering on VQA v2. All reported metrics, AP \% and AF \%, are measured under the COMP setting on the G1 split. The Mean $\pm$ Standard Deviation ($\sigma$) is calculated across the three permutation orders.}
\label{tab:task_order_results}
\resizebox{0.9\columnwidth}{!}{
\begin{tabular}{l cccc}
\toprule
\multirow{2}{*}{\textbf{Task Order}} & \multicolumn{2}{c}{\textbf{Vanilla}} & \multicolumn{2}{c}{\textbf{AIM (Ours)}} \\
\cmidrule(lr){2-3} \cmidrule(lr){4-5} 
 & \textbf{AP $\uparrow$} & \textbf{AF $\downarrow$} & \textbf{AP $\uparrow$} & \textbf{AF $\downarrow$} \\
\midrule
Default Order & 11.79 & 27.16 & 41.92 & 3.12 \\
Reverse Order & 5.21  & 33.75 & 41.22 & 0.58 \\
Random Order  & 2.70  & 33.39 & 40.02 & 2.90 \\
\midrule
\textbf{Mean $\pm$ $\sigma$} & 6.57 $\pm$ 4.69 & 31.43 $\pm$ 3.71 & \textbf{41.05 $\pm$ 0.96} & \textbf{2.20 $\pm$ 1.41} \\
\bottomrule
\end{tabular}
}
\end{table}

\subsection{Parameter Sensitivity Analysis}
\label{supp:parameter_sensitivity}

To investigate the impact of the asymmetric masking strategy, we conduct a parameter sensitivity analysis on the masking ratios. Fixing the shared modality masking ratio at $\rho_{shared}=0.1$, we perform a grid search over the visual masking ratio $\rho_{vis} \in \{0.1, 0.3, 0.5\}$ and the text masking ratio $\rho_{text} \in \{0.3, 0.5, 0.7\}$. The model is evaluated under the COMP setting on the G1 split using an episodic memory buffer of $M=500$.

The AP and AF across both Novel and Seen compositions are reported in Table~\ref{tab:param_sensitivity}. The results demonstrate a plasticity-stability trade-off. Increasing the text masking ratio from $\rho_{text}=0.3$ to $\rho_{text}=0.5$ yields improvements in AP across most visual masking settings, supporting the hypothesis that the text decoder requires stronger regularization to mitigate catastrophic forgetting. However, while further increasing the text sparsity to $\rho_{text}=0.7$ achieves the lowest forgetting rates (e.g., AF decreases to 3.59\% at $\rho_{vis}=0.1$), it corresponds with a degradation in AP. This observation suggests that high text masking ($\rho_{text} \ge 0.7$) may induce a representation bottleneck, limiting the model's capacity to encode complex compositional concepts. Consequently, the asymmetric configuration of ($\rho_{vis}=0.3, \rho_{text}=0.5$) provides a favorable balance, achieving competitive performance (Novel AP: 38.89\%, Seen AP: 39.63\%) while effectively mitigating forgetting.

\begin{table*}[t]
\centering
\caption{Parameter sensitivity analysis of the masking ratios $\rho_{vis}$ and $\rho_{text}$ under the COMP setting on the G1 split ($M=500$). The shared ratio is fixed at $\rho_{shared}=0.1$. Each cell reports \textbf{AP / AF} (\%).}
\label{tab:param_sensitivity}
\resizebox{0.8\textwidth}{!}{ 
\begin{tabular}{l | ccc | ccc}
\toprule
\multirow{2}{*}{$\rho_{vis}$} & \multicolumn{3}{c|}{\textbf{Novel (AP $\uparrow$ / AF $\downarrow$)}} & \multicolumn{3}{c}{\textbf{Seen (AP $\uparrow$ / AF $\downarrow$)}} \\
\cmidrule(lr){2-4} \cmidrule(lr){5-7} 
 & $\rho_{text}=0.3$ & $\rho_{text}=0.5$ & $\rho_{text}=0.7$ & $\rho_{text}=0.3$ & $\rho_{text}=0.5$ & $\rho_{text}=0.7$ \\
\midrule
$0.1$ & 37.08 / 5.68 & 38.09 / 4.78 & 37.52 / \textbf{3.59} & 38.47 / 6.10 & \textbf{39.95} / 5.09 & 38.13 / \textbf{3.69} \\
$0.3$ & 37.29 / 6.48 & \textbf{38.89} / 5.31 & 36.97 / 4.12 & 38.81 / 6.18 & 39.63 / 5.29 & 37.60 / 4.63 \\
$0.5$ & 37.86 / 5.37 & \textbf{38.89} / 4.45 & 37.25 / 3.94 & 38.40 / 6.15 & 39.01 / 6.03 & 38.06 / 4.31 \\
\bottomrule
\end{tabular}
}
\end{table*}

\section{Deep Dive into Asymmetry Dynamics}
\label{supp:param_shift}

\subsection{Empirical Analysis of Parameter Dynamics}
This section provides the numerical data supporting the empirical analysis of parameter dynamics presented in Section 4 (Figure 2). Table~\ref{tab:empirical_shift} reports the parameter capacity, $L_2$ parameter shift, and Mean Fisher Information for each modality-specific subspace. To quantify the interference across tasks, we compute the \textit{Weighted Shift}, which integrates parameter variation with its corresponding sensitivity.

Under the \textit{Vanilla} fine-tuning baseline, the text subspace ($\Theta_{text}$) exhibits a maximum $L_2$ shift of 3.38\%. Due to its large parameter volume (166.32M), this shift results in a proportionally large weighted penalty, which contributes to the degradation of previously acquired reasoning capabilities. Concurrently, the shared subspace ($\Theta_{shared}$) records the highest parameter drift, reaching an 18.82\% $L_2$ shift. The vision projector ($\Theta_{vis}$) also undergoes continuous updates despite exhibiting the highest Mean Fisher sensitivity among the three modules. These quantitative measurements indicate a gradient imbalance inherent in symmetric optimization, supporting the rationale for the proposed asymmetric protection strategy.

\begin{table*}[t]
\centering
\caption{Detailed parameter shift and sensitivity values for the Vanilla baseline across the 10 VQA v2 tasks. This table provides the exact $L_2$ Shift (\%), Mean Fisher, and Weighted Shift used for the analysis in Figure 2. Scientific notation is used for extreme values.}
\label{tab:empirical_shift}
\resizebox{\textwidth}{!}{
\begin{tabular}{ll cccccccccc}
\toprule
\textbf{Subspace} & \textbf{Metric} & \textbf{Rec.} & \textbf{Loc.} & \textbf{Jud.} & \textbf{Com.} & \textbf{Cou.} & \textbf{Act.} & \textbf{Col.} & \textbf{Typ.} & \textbf{Sub.} & \textbf{Cau.} \\
\midrule
\multirow{3}{*}{\textbf{Vision ($\Theta_{vis}$, 1.58M)}} 
& $L_2$ Shift (\%) & 1.26 & 0.18 & 0.78 & 0.18 & 0.65 & 0.37 & 0.68 & 0.30 & 0.32 & 0.11 \\
& Mean Fisher & 5.8e-10 & 6.3e-09 & 7.9e-11 & 1.9e-11 & 8.7e-10 & 2.0e-09 & 4.0e-09 & 1.8e-09 & 9.1e-10 & 2.1e-10 \\
& Weighted Shift & 3.2e-07 & 3.3e-08 & 6.9e-09 & 1.3e-10 & 4.9e-08 & 5.7e-08 & 3.1e-07 & 1.9e-08 & 1.3e-08 & 5.5e-10  \\
\midrule
\multirow{3}{*}{\textbf{Text ($\Theta_{text}$, 166.32M)}} 
& $L_2$ Shift (\%) & 3.38 & 0.46 & 1.51 & 0.39 & 0.93 & 0.76 & 0.78 & 0.78 & 0.74 & 0.32  \\
& Mean Fisher & 1.1e-09 & 4.4e-09 & 2.6e-10 & 4.2e-11 & 1.2e-10 & 1.0e-09 & 3.3e-10 & 1.7e-09 & 2.0e-10 & 7.0e-10  \\
& Weighted Shift & 3.7e-05 & 3.3e-06 & 2.1e-06 & 1.3e-08 & 2.1e-06 & 2.4e-06 & 2.1e-06 & 3.6e-06 & 4.1e-07 & 3.1e-07 \\
\midrule
\multirow{3}{*}{\textbf{Shared ($\Theta_{shared}$, 57.82M)}} 
& $L_2$ Shift (\%) & 18.82 & 5.08 & 17.12 & 5.59 & 10.38 & 8.66 & 10.92 & 7.80 & 8.22 & 3.43  \\
& Mean Fisher & 1.8e-08 & 9.2e-08 & 4.0e-09 & 6.7e-10 & 1.0e-08 & 2.9e-08 & 4.2e-08 & 6.5e-08 & 1.8e-08 & 4.2e-09  \\
& Weighted Shift & 4.8e-05 & 2.0e-05 & 1.2e-05 & 2.8e-07 & 1.3e-05 & 1.7e-05 & 5.4e-05 & 3.3e-05 & 1.1e-05 & 5.4e-07  \\
\bottomrule
\end{tabular}
}
\end{table*}

\subsection{Parameter Sensitivity and Layer-wise Dynamics}
\label{supp:param_sensitivity}

\begin{figure*}[t]
    \centering
    \includegraphics[width=\textwidth]{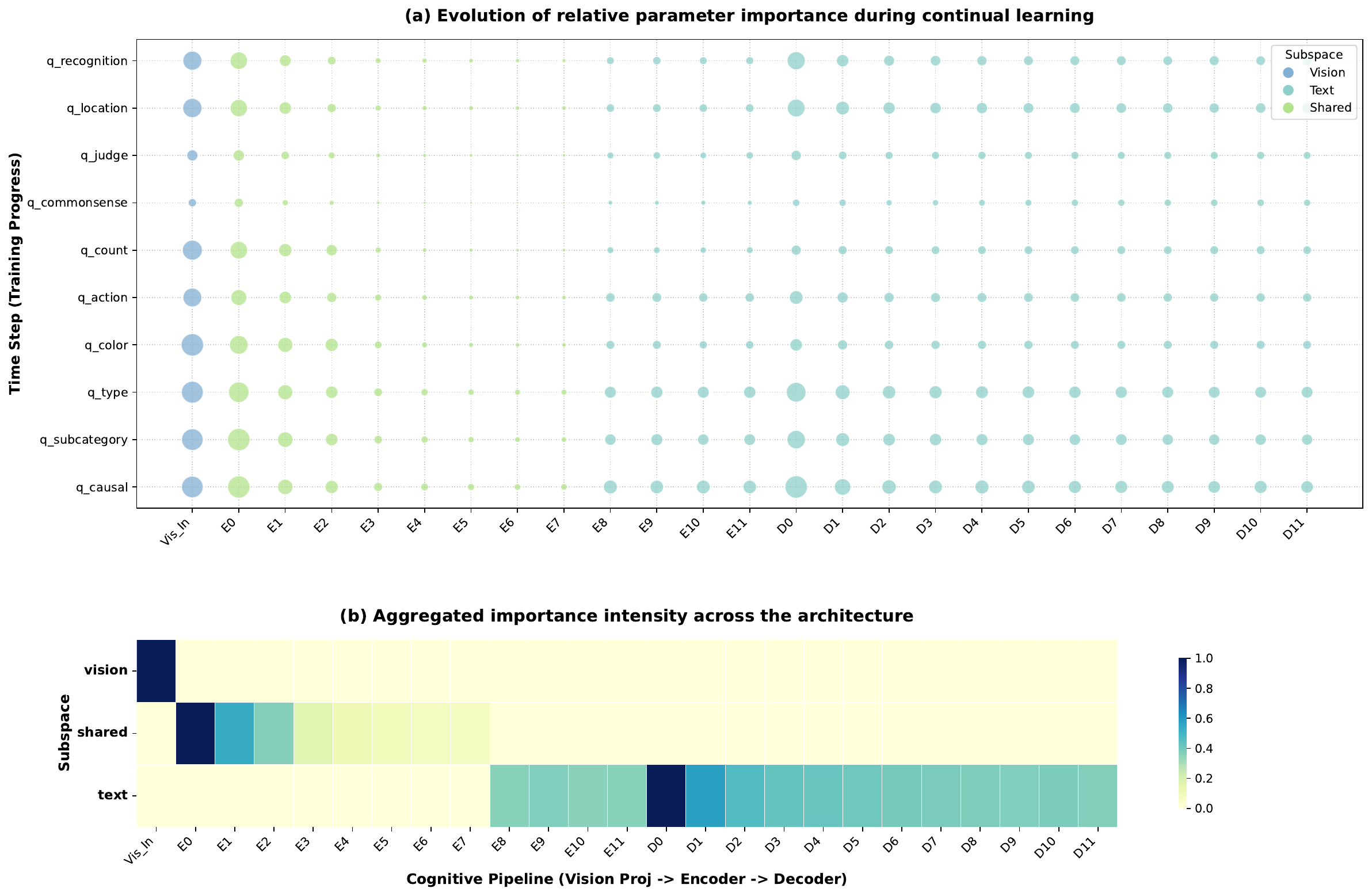}
    \caption{Dynamic analysis of parameter sensitivity across the VL-T5 architecture during continual learning. The x-axis represents 25 functional nodes from the visual projector to the final decoder layer. \textbf{(a)} The temporal evolution of relative parameter importance for the Vision, Text, and Shared subspaces across 10 tasks. \textbf{(b)} The aggregated importance intensity heatmap, indicating that multimodal alignment is concentrated in shallow layers while semantic reasoning localizes in deep layers.}
    \label{fig:fisher_dynamics}
\end{figure*}

To analyze the mechanics of the Asymmetric Information Masking (AIM) framework, we track the parameter sensitivity dynamics throughout the continual learning process. Figure~\ref{fig:fisher_dynamics} illustrates the relative Fisher Information across the network architecture, from the visual projection layer (\textit{Vis\_In}) through the encoder (\textit{E0-E11}) and decoder (\textit{D0-D11}) blocks.

\paragraph{Layer-wise Modality Distribution.} 
The aggregated heatmap (Figure~\ref{fig:fisher_dynamics}b) indicates a distinct spatial distribution of parameter importance. The \textit{Vision} subspace is confined to the visual projection layer. The \textit{Shared} subspace exhibits a bottom-heavy distribution; its Mean Fisher values peak in the shallow encoder layers (\textit{E0-E2}) and decay across subsequent layers (\textit{E3-E7}). This measurement suggests that foundational multimodal alignment (e.g., mapping visual features to the semantic space) primarily occurs in the early stages of the encoder. Conversely, the \textit{Text} subspace accounts for the majority of parameter sensitivity in the deep encoder layers (\textit{E8-E11}) and the entire decoder (\textit{D0-D11}), which correspond to higher-level semantic reasoning and text generation.

\paragraph{Temporal Evolution Across Tasks.} 
Figure~\ref{fig:fisher_dynamics}a tracks these sensitivities across the 10 sequential tasks. The relative distribution of parameter importance across the three subspaces remains consistent regardless of the specific task being learned. The \textit{Shared} parameters maintain high Fisher values in the shallow layers across all time steps, stabilizing the visual-linguistic mapping. Concurrently, the \textit{Text} parameters exhibit localized sensitivity fluctuations within the deep decoder layers, reflecting parameter updates required for task-specific adaptation.

These observations support the rationale for asymmetric regularization. Applying a uniform constraint across the architecture introduces a trade-off: it may either over-regularize the deep layers, restricting semantic plasticity, or under-regularize the shallow layers, disrupting multimodal alignment. The AIM formulation is structurally designed to accommodate this imbalance.

\section{Efficiency and Computational Overhead}
\label{supp:efficiency}

Since the proposed Asymmetric Information Masking (AIM) framework maintains the original backbone architecture of the underlying MLLM without introducing auxiliary networks, its base training complexity aligns with standard fine-tuning. The additional computational overhead arises exclusively from computing the empirical Fisher Information Matrix (FIM) and generating the binary masks. 

In our implementation, AIM calculates the FIM diagonal once at the end of each task using a randomly sampled subset of $N=500$ instances. We profile the system overhead based on training logs across three dimensions, as summarized in Table~\ref{tab:efficiency}:

\begin{itemize}
    \item \textbf{Computational Volume:} FIM computation requires one additional forward and backward pass per sampled instance. For a representative continual learning task (e.g., \textit{q\_judge}), the model processes approximately \textbf{303,000} training instances (across 3,792 batches). Processing the additional 500 instances for FIM results in an approximately \textbf{0.16\%} increase in the total optimization volume.
    \item \textbf{Wall-clock Time:} The one-time FIM and masking computation incurs a fixed overhead of 60.7 seconds per task. Compared to the standard training duration of the representative task (43.80 minutes), this introduces a relative time overhead of \textbf{2.3\%}.
    \item \textbf{GPU Memory (VRAM) Footprint:} AIM requires caching the binary parameter masks and the FIM diagonal. For the 224.5M parameters of our architecture, storing these static tensors consumes 428.28 MB of additional GPU memory. During training, peak VRAM usage increases from 11.04 GB to \textbf{11.46 GB}, yielding a \textbf{3.8\%} relative footprint increment.
\end{itemize}

\begin{table}[t]
\centering
\caption{System overhead profiling of AIM. The metrics are measured against the standard rehearsal baseline on a representative task (\textit{q\_judge}).}
\label{tab:efficiency}
\resizebox{\columnwidth}{!}{
\begin{tabular}{l ccc}
\toprule
\textbf{Metric} & \textbf{Baseline} & \textbf{AIM (Ours)} & \textbf{Overhead} \\
\midrule
Opt. Volume & 303.3K & 303.8K (+500) & + 0.16\% \\
Time / Task & 43.8 min & 44.8 min (+60s) & + 2.3\% \\
Peak VRAM   & 11.0 GB & 11.4 GB (+428MB)& + 3.8\% \\
\bottomrule
\end{tabular}
}
\end{table}

\begin{figure*}[t]
  \centering
  \includegraphics[width=0.9\textwidth]{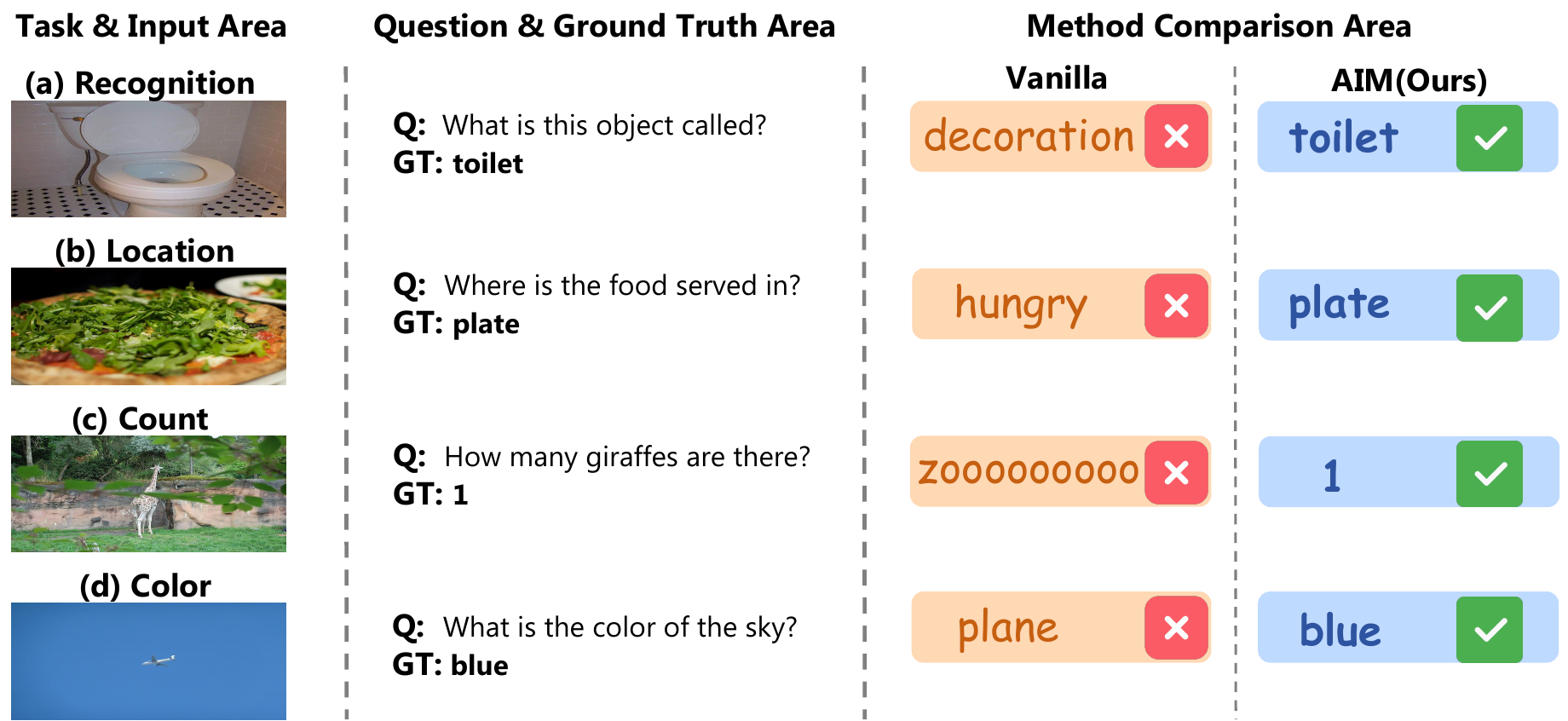}
  \caption{Qualitative examples demonstrating catastrophic forgetting in the Vanilla baseline versus knowledge retention in AIM.}
  \label{fig:supp_cases}
\end{figure*}

\section{Qualitative Analysis}
\label{supp:qualitative}

To provide a granular view of the catastrophic forgetting phenomenon, Figure~\ref{fig:supp_cases} presents a qualitative comparison between the \textit{Vanilla} baseline and our AIM framework. We sample representative cases across distinct reasoning dimensions, including recognition, location, counting, and color.

After sequential training, the Vanilla baseline frequently exhibits semantic drift and a loss of visual grounding. Instead of producing accurate answers based on the given image, the model tends to output generic tokens or task-irrelevant concepts. For instance, it may output an abstract or unrelated word for a precise recognition query, or fail to ground its response for a spatial location task. These failure modes indicate that standard symmetric updates compromise previously established visual-linguistic alignments, forcing the model to rely on biased language priors rather than true cross-modal reasoning.

In contrast, AIM successfully maintains accurate and visually grounded predictions across the same diverse set of sample queries. By yielding correct responses on earlier concepts after sequential adaptation, AIM demonstrates the effectiveness of its asymmetric isolation mechanism. These results further validate our core finding: applying modality-specific masking protects the sensitive visual bottleneck from optimization interference, thereby preserving the structural integrity of the reasoning pathways throughout the continual learning process.

\end{document}